\pdfoutput=1
\documentclass[final,onefignum,onetabnum,hidelinks]{siamart251216}
\usepackage[utf8]{inputenc}
\usepackage{microtype,amssymb,bm,mathtools}
\usepackage{etoolbox}
\usepackage{graphicx,booktabs,array,tabularx,adjustbox}
\usepackage[compatibility=false]{caption}
\usepackage{siunitx,placeins,url}
\usepackage{multirow,enumitem}
\usepackage[algo2e,ruled,vlined]{algorithm2e}
\providecommand{\widebar}[1]{\overline{#1}}
\DeclareSIUnit{\psu}{psu}
\DeclareUnicodeCharacter{02BC}{'}
\crefname{algocf}{Algorithm}{Algorithms}
\Crefname{algocf}{Algorithm}{Algorithms}
\crefname{appendix}{Appendix}{Appendices}
\Crefname{appendix}{Appendix}{Appendices}
\newcommand{\vect}[1]{\mathbf{#1}}
\newcommand{\mat}[1]{\mathbf{#1}}
\newcommand{\ZeroVec}[1]{\mathbf{0}_{#1}}
\newcommand{\Rdim}[1]{\mathbb{R}^{#1}}
\newcommand{\rv}[1]{\mathsf{#1}}
\newcommand{\E}{\mathbb{E}}
\newcommand{\dd}{\mathrm{d}}
\newcommand{\Normal}[2]{\mathcal{N}\!\left(#1,\,#2\right)}
\newcommand{\Unif}[1]{\mathrm{Uniform}\left(#1 \right)}
\newcommand{\BetaDist}[2]{\mathrm{Beta}\left(#1,\,#2 \right)}
\newcommand{\DKL}[2]{\mathcal{D}_{\mathrm{KL}}\!\left(#1 \| #2 \right)}
\newcommand{\param}{\theta}
\newcommand{\ctx}{u}
\newcommand{\obs}{y}
\newcommand{\ctxi}[1]{\ctx_{#1}}
\newcommand{\Param}{\bm{\theta}}
\newcommand{\Ctx}{\vect{\ctx}}
\newcommand{\Obs}{\vect{\obs}}
\newcommand{\ParamRef}{\bm{\theta}^{\mathrm{Ref}}}
\newcommand{\ParamSpace}{\Theta}
\newcommand{\CtxSpace}{\mathcal{U}}
\newcommand{\ObsSpace}{\mathcal{Y}}
\newcommand{\Ctxdim}{d_{\ctx}}
\newcommand{\Obsdim}{d_{\obs}}
\newcommand{\Ctxi}[1]{\Ctx_{#1}}
\newcommand{\Obsi}[1]{\Obs_{#1}}
\newcommand{\CtxTruei}[1]{\Ctx^\mathrm{True}_{#1}}
\newcommand{\Ctxn}{\Ctx_{1:n}}
\newcommand{\Obsn}{\Obs_{1:n}}
\newcommand{\PCtx}{P_{\mathrm{ctx}}}
\newcommand{\rvCtxNoiseCov}{\mat{Q}}
\newcommand{\rvObsNoiseCov}{\mat{R}}
\newcommand{\Dataset}{\mathcal{D}}
\newcommand{\rvParam}{{\bm{\Theta}}}
\newcommand{\rvCtx}{\rv{U}}
\newcommand{\rvObs}{\rv{Y}}
\newcommand{\rvCtxi}[1]{\rvCtx_{#1}}
\newcommand{\rvObsi}[1]{\rvObs_{#1}}
\newcommand{\rvCtxn}{\rvCtx_{1:n}}
\newcommand{\ModelObs}{\mathcal{G}}
\newcommand{\ModelObsPhys}{\ModelObs_\mathrm{Phys}}
\newcommand{\ModelObsCall}[2]{\ModelObs\!\left(#1,\,#2\right)}
\newcommand{\ParamOpt}{\Param^*}
\newcommand{\VarParam}{\phi}
\newcommand{\qvar}{q_{\VarParam}}
\newcommand{\qvarof}[1]{q_{\VarParam}(#1)}
\newcommand{\FlowMapName}{\mathcal{T}}
\newcommand{\rvBase}{\rv{Z}}
\newcommand{\BaseLatent}{q_{\rvBase}}
\newcommand{\FlowMapVI}{\FlowMapName_{\VarParam}}
\newcommand{\FlowMapVIInv}{\FlowMapVI^{-1}}
\newcommand{\CtxStar}{\Ctx^\star}
\newcommand{\LocalPost}{\pi_{\CtxStar}}
\newcommand{\LocalZof}[3]{Z\left(#1, #2, #3 \right)}
\newcommand{\Bandwidth}{h}
\newcommand{\Kernelh}{K_{\Bandwidth}}
\newcommand{\BandwidthMat}{\mat{H}}
\newcommand{\ScaleCtxi}[1]{\sigma_{#1}}
\newcommand{\CtxScaleMat}{\Sigma}
\newcommand{\AlgWeighti}[1]{w_{#1}}
\newcommand{\Weighti}[1]{w_{#1}(\Bandwidth(\CtxStar);\, \CtxStar,\, \Ctxn)}
\newcommand{\ScalingFactor}{\tau}
\newcommand{\TauLoc}[3]{\ScalingFactor\left(#1;\, #2,\,#3\right)}
\newcommand{\ESS}{\mathrm{ESS}_{\mathrm{Kish}}}
\newcommand{\ESSLoc}[3]{\ESS\left(#1;\, #2,\,#3\right)}
\newcommand{\LocRateTarget}{\ell_\star}
\newcommand{\LocRate}[3]{\ell\left(#1;\, #2,\,#3\right)}
\newcommand{\hmin}{\Bandwidth_{\min}}
\newcommand{\hmax}{\Bandwidth_{\max}}
\newcommand{\LocELBO}{\mathrm{LocELBO}}
\newcommand{\LocELBOof}[4]{\LocELBO(#1; #2, #3, #4)}
\newcommand{\AlgLocNELBOof}[2]{\mathcal{L}(#1; #2)}
\newcommand{\AlgLocNELBOMCof}[2]{\hat{\mathcal{L}}(#1; #2)}
\newcommand{\TransWidth}{\omega}
\newcommand{\nepoch}{n_{\mathrm{epoch}}}
\newcommand{\btarget}{b_{\mathrm{target}}}
\newcommand{\bobs}{b_{\mathrm{obs}}}
\newcommand{\nparticle}{n_{\mathrm{particle}}}
\newcommand{\bracketmult}{c}
\newcommand{\bracketexp}{\beta}
\newcommand{\Temp}{T}
\newcommand{\Sal}{S}
\newcommand{\ZonalVel}{v}
\newcommand{\Depth}{\zeta}
\newcommand{\ViscVert}{K_{\mathrm{M}}}
\newcommand{\DiffVert}{K_{\mathrm{T}}}
\newcommand{\Distorsion}{\Delta}
\newcommand{\HeatForcing}{\mathcal{Q}_{\mathrm{sfc}}}
\newcommand{\WindForcing}{\tau_{\mathrm{sfc}}}
\newcommand{\ParamCmu}{c_{\mu0}}
\newcommand{\ParamCeps}{c_{\varepsilon 1}}
\newcommand{\MLD}{\Depth_\mathrm{MLD}}
\newcommand{\DepthMax}{H}
\newcommand{\NDepth}{N_{\Depth}}
\newcommand{\DepthControl}{\Depth_{\mathrm{c}}}
\newcommand{\SimDuration}{T_{\mathrm{sim}}}
\newcommand{\TimeStep}{\Delta t}
\newcommand{\OutputTimeStep}{\Delta t_{\mathrm{out}}}
\newcommand{\RefDensity}{\rho_0}
\newcommand{\Gravity}{g}
\newcommand{\HeatCapacity}{c_p}
\newcommand{\ThermalExpansion}{\alpha}
\newcommand{\HalineContraction}{\beta}
\newcommand{\VonKarman}{\kappa}
\newcommand{\Coriolis}{f}
\newcommand{\CalibSize}{n_{\mathrm{calib}}}
\newcommand{\TestSize}{n_{\mathrm{test}}}
\newcommand{\TargetSize}{n_{\mathrm{target}}}
\newcommand{\ShearProduction}{P_s}
\newcommand{\BuoyancyProduction}{P_b}
\newcommand{\DissipationRate}{\varepsilon}
\newcommand{\Density}{\rho}
\newcommand{\ParamTrue}{\Param^\mathrm{Truth}}
\newcommand{\RefObsi}[1]{\Obs_{#1}^{\star}}
\newcommand{\npred}{n_{\mathrm{pred}}}
\newcommand{\ntest}{n_{\mathrm{test}}}
\newcommand{\Error}{e}

\newcommand{\ErrorMean}{\bar{\Error}}
\newcommand{\ErrorStd}{\mathrm{Std}(\Error)}

\newcommand{\ErrorUpperTailMean}[1]{\overline{\Error}_{>#1}}
\hypersetup{hidelinks,unicode,pdfauthor={Exauc\'e Luweh Adjim Ngarti, Elise Arnaud, L\'eo Nicoletti, Victor Trappler, Arthur Vidard}}

\makeatletter

  \renewcommand\section{\@startsection{section}{1}{.25in}%
    {1.3ex \@plus .5ex \@minus .2ex}{.6ex \@plus .1ex}%
    {\reset@font\normalsize\bfseries}}
  \renewcommand\subsection{\@startsection{subsection}{2}{.25in}%
    {1.3ex \@plus .5ex \@minus .2ex}{.6ex \@plus .1ex}%
    {\reset@font\normalsize\bfseries}}
  \renewcommand\append[2][?]{%
    \refstepcounter{section}%
    \par\addvspace{\@sectionspace}%
    {\reset@font\normalsize\bfseries\appendixname\ \thesection. #2\@mypunct{}\par}%
    \addcontentsline{toc}{section}{\appendixname~\thesection. #2}%
    \nobreak\vspace{.6ex}\@afterheading}
\patchcmd{\thebibliography}{\sloppy}{\setlength{\itemsep}{0pt}\sloppy}{}{}
\makeatother
\title{Amortized Localized Posterior Inference for Model Calibration
with Application to an Ocean Turbulence Closure}

\newcommand{\paperauthors}{%
  Exauc\'e Luweh Adjim Ngarti\textsuperscript{1,2}\textsuperscript{*},
  Elise Arnaud\textsuperscript{1},
  L\'eo Nicoletti\textsuperscript{2},\\[0.35em]
  Victor Trappler\textsuperscript{3},
  Arthur Vidard\textsuperscript{1}%
}

\author{\paperauthors\thanks{%
Corresponding author: Exauc\'e Luweh Adjim Ngarti.
Email: \href{mailto:exaucengarti@gmail.com}
{\nolinkurl{exaucengarti@gmail.com}}.}}
\headers{}{}

\newcommand{\paperaffiliations}{%
  \begingroup
  \centering
  \textsuperscript{1} Univ. Grenoble Alpes, Inria, CNRS,
  LJK, 38000 Grenoble, France.\par
  \textsuperscript{2} Bull,
Centre of Excellence in Performance Programming (CEPP),
France.\par
  \textsuperscript{3} Department GMI, Institut Henri Fayol,
  Mines Saint-\'Etienne, Univ Clermont Auvergne,
  CNRS, UMR 6158 LIMOS,
  F-42023 Saint-\'Etienne, France.\par
  \endgroup
}

\date{}

\headers{}
        {}

\makeatletter
\let\paper@thanks\thanks
\renewcommand{\thanks}[1]{\begingroup\let\@makefnmark\relax\paper@thanks{#1}\endgroup}
\makeatother

\begin{document}
\maketitle
\paperaffiliations
\vspace{1em}
\begin{abstract}
Model discrepancy may prevent a single calibration parameter from reproducing a physical response across the context domain. In addition, uncertainties in observations and contexts affect calibration and must be accounted for and quantified. We construct a family of localized posteriors by weighting observation likelihoods according to their proximity to a target context. Gaussian localization weights incorporate context uncertainty through covariance inflation, with reflection at domain boundaries and an optional correction for uneven context sampling. A conditional normalizing flow approximates the family of posteriors through amortized variational inference. Integrating the localized posteriors under a prescribed context measure gives the marginalized parameter distribution. In a twin experiment with the Branin--Hoo model, we compare global and localized inference under uniform and skewed context distributions. Localization reduces the reported aggregate predictive errors relative to the global posterior. An application to the $k$--$\varepsilon$ turbulence closure in a single-column ocean model, with atmospheric surface forcing as context, shows that localized posterior calibration reduces temperature and zonal-velocity errors relative to deterministic and global-posterior calibration. Both twin experiments are constructed to admit local compensation.
\end{abstract}

\section{Introduction}\label{sec:introduction}

Numerical models play a central role in science and engineering. In geosciences, they form the basis of weather forecasting \cite{bauer_quiet_2015}, climate modeling \cite{heavens2013studying,alizadeh_advances_2022}, hydrological simulation \cite{fatichi_overview_2016}, and ocean dynamics \cite{griffies_fundamentals_2018}.  Their construction typically starts from fundamental equations, such as the Navier--Stokes equations in fluid dynamics, and proceeds through modeling assumptions, approximations, and numerical discretizations.

However, numerical models are not exact representations of the physical world. Many physical processes occur at spatial or temporal scales that cannot be explicitly resolved at an affordable computational cost. Their effects must therefore be represented indirectly through simplified closure laws, parameterizations, or empirical corrections. As a result, numerical models involve parameters that control unresolved or imperfectly known mechanisms. These parameters cannot usually be determined from the governing equations alone and must instead be estimated from observational data. This estimation process is referred to as \emph{model calibration}.

In practice, calibration compares model outputs with observations collected under different physical or experimental conditions, referred to here as \emph{contexts}. A common approach is to estimate a single calibration parameter that explains the observations across the context domain. However, such a parameter may not exist when missing or simplified mechanisms induce model discrepancy that varies with context. Moreover, calibration data may over-represent certain contexts, while the contexts associated with observations may themselves be uncertain. These effects can influence the inferred parameters and the predictive performance of the calibrated model across the context domain.

One approach to model discrepancy is to augment the numerical response, for example with a Gaussian process or a neural network \cite{kennedy_bayesian_2001,brynjarsdottir2014learning,ling2014selection,leoni2024bayesian,yang2025active}. Such approaches increase modeling flexibility but introduce additional unknowns and may complicate physical interpretation. Calibration under uncertain or unobserved contexts has also been considered \cite{trappler2021controle,radivsic2025consideration}. Here, we consider contexts observed with uncertainty and address model discrepancy by allowing the calibration distribution to depend on the target context, without introducing an explicit discrepancy model.

Our approach relies on a local compensability assumption: although a single parameter may not adequately reproduce observations across the entire context domain, model discrepancy can be approximately compensated by calibration parameters within a neighborhood of a given context. We therefore construct a family of localized posterior distributions, one for each target context, by weighting the observation likelihood contributions according to their proximity to that context. Observations collected near the target context receive greater weight, allowing the calibration distribution to vary across the context domain.

Since the recorded contexts may be uncertain, we incorporate context uncertainty into Gaussian localization weights through covariance inflation. We also account for domain boundaries through reflection and consider an optional correction for uneven context sampling. The resulting localized posteriors describe parameter uncertainty at each target context. Integrating these distributions under a prescribed context measure yields a marginalized parameter distribution, which summarizes the context-dependent calibration uncertainty.

Constructing a separate posterior approximation for every target context would require repeated inference. To avoid this, we use conditional normalizing flows \cite{rezende2015variational,dinh2016density} and amortized variational inference \cite{jordan1999introduction,hoffman2013stochastic,blei2017variational}. A single conditional flow is trained over the context domain to approximate the family of localized posteriors, allowing a posterior approximation to be evaluated at a new target context without repeating the optimization.

We first examine the method in a Branin--Hoo twin experiment. We compare global and localized posterior inference under uniform and skewed context distributions, in a setting where local compensation is possible by construction. Localization reduces aggregate predictive errors relative to global posterior calibration.

We then apply the approach to the calibration of a $k$--$\varepsilon$ turbulence closure in a single-column ocean model. Ocean models represent unresolved turbulent mixing through closure schemes whose parameters must be specified. Vertical mixing redistributes momentum and tracers under the combined effects of stratification, surface heat flux, and wind-driven shear. We consider temperature and zonal-velocity observations, with atmospheric surface forcing, including wind stress and heat flux, defining the context. These forcing quantities are uncertain, notably because they may be estimated from bulk atmospheric variables \cite{brunke2011assessment}. Moreover, the simplified turbulence closure may introduce context-dependent model discrepancy, making a single calibration parameter inadequate across different forcing conditions. In the designed experiments, localized posterior calibration reduces temperature and zonal-velocity errors relative to deterministic and global-posterior calibration. The results also show reduced parameter identifiability under weak-wind and heating conditions.

The main contributions of this work are:
\begin{itemize}
    \item A localized Bayesian calibration formulation based on a local compensability assumption, in which observation likelihood contributions are weighted according to their proximity to a target context.

    \item A localization procedure accounting for uncertainty in the observed contexts, domain boundaries, and uneven context sampling, together with a marginalized parameter distribution under a prescribed context measure.

    \item An amortized variational inference approach using conditional normalizing flows to approximate the family of localized posteriors over the context domain.

    \item A numerical assessment using a Branin--Hoo twin experiment and the calibration of an ocean turbulence closure, comparing localized inference with global calibration and examining predictive errors and parameter identifiability.
\end{itemize}

\section{Problem setting}\label{sec:problem_setting}

We consider the calibration of a numerical model using observations
collected under different physical contexts.

\subsection{Model discrepancy and context}
\label{sec:discrepancy_context}

Let $\ModelObs:\ParamSpace\times\CtxSpace\to\ObsSpace$
denote the numerical model, $\Param\in\ParamSpace$ the model
parameter to be calibrated, and $\Ctx\in\CtxSpace$ the context
variable specified for each model evaluation.
Let $\ModelObsPhys(\Ctx)$ denote the physical response at
context $\Ctx$. Following \cite{kennedy_bayesian_2001},
we write
\begin{equation}\label{eq:model_discrepancy}
    \ModelObsPhys(\Ctx)
    =
    \ModelObsCall{\ParamOpt}{\Ctx}
    +
    \delta(\Ctx),
\end{equation}
where $\ParamOpt$ is a reference parameter and $\delta$
represents the context-dependent discrepancy between the
numerical model and the physical response.

Because this discrepancy may vary with context, there is
no guarantee that a single parameter $\Param^*$ satisfies
\[
    \ModelObsCall{\Param^*}{\Ctx}
    =
    \ModelObsPhys(\Ctx)
\]
for every $\Ctx\in\CtxSpace$.
We instead assume local compensability: for each target
context, a suitable calibration parameter can approximately
compensate for the discrepancy in its neighborhood.
We further assume that adequate parameters at nearby
contexts are sufficiently similar for nearby observations
to be informative.

\subsection{Observation and context uncertainty}
\label{sec:observation_uncertainty}

Let
$\Dataset_n=\{(\Ctxi{i},\Obsi{i})\}_{i=1}^n$
denote the calibration dataset, consisting of observations
and their associated recorded contexts.
We assume conditionally independent observations with
additive Gaussian errors, giving the working likelihood
\begin{equation}\label{eq:observation_likelihood}
    \rvObsi{i}
    \mid \Param,\Ctxi{i}
    \sim
    \Normal{
        \ModelObsCall{\Param}{\Ctxi{i}}
    }{\rvObsNoiseCov},
    \qquad i=1,\ldots,n,
\end{equation}
where
$\rvObsNoiseCov\in\Rdim{\Obsdim\times\Obsdim}$
is a symmetric positive-definite covariance matrix.

We assign a prior distribution to the calibration parameter,
independent of the recorded contexts:
\begin{align}
\label{eq:param_independent_of_context}
    p_{\rvParam \mid \rvCtxn}
    \left(
        \Param \mid \Ctxn
    \right)
    &=
    p_{\rvParam}
    \left(
        \Param
    \right).
\end{align}

In addition to observation uncertainty, the recorded
contexts are affected by additive Gaussian errors.
Writing $\Ctxi{i}^{\mathrm{true}}$ for the true context
associated with observation $i$, we assume
\begin{equation}\label{eq:context_observation_model}
    \Ctxi{i}
    =
    \Ctxi{i}^{\mathrm{true}}
    +
    \varepsilon_i^{(c)},
    \qquad
    \varepsilon_i^{(c)}
    \sim
    \Normal{\ZeroVec{\Ctxdim}}{\rvCtxNoiseCov},
\end{equation}
where $\rvCtxNoiseCov$ is the known context-error covariance.
The context errors are assumed independent across observations
and independent of the observation errors.

The working likelihood in \cref{eq:observation_likelihood}
is evaluated at the recorded contexts. Their uncertainty
is incorporated into the localization weights introduced
below, rather than by explicitly integrating the numerical
model over the unknown true contexts.

\subsection{Calibration objectives}
\label{sec:calibration_objectives}

Our objective is to infer parameter distributions that
provide adequate local calibration across the context
domain while accounting for observation and context
uncertainties and uneven context sampling.

We also seek a parameter distribution relevant to a
prescribed probability measure $\PCtx$ on $\CtxSpace$.
This measure identifies the contexts over which calibration
performance is of interest and need not coincide with
their empirical distribution in the dataset.

These objectives motivate a family of context-dependent
posterior distributions, whose approximation and aggregation
are developed in the next section.

\section{Amortized localized posterior inference}
\label{sec:amortized_localized_inference}

Under the local compensability assumption, observations collected
near a target context are expected to be more informative for
calibration at that context. We use this principle to construct
a family of localized posterior distributions. We first define
the localized likelihood and its weights, including the treatment
of context uncertainty and boundaries. We then control the
localization strength through an effective sample size, approximate
the posterior family using a conditional normalizing flow, and
aggregate the localized posteriors under the prescribed context
measure.

\subsection{Localized posterior}
\label{sec:localized_posterior}

Following the localized likelihood principle
\cite{tibshirani1987local}, we assign greater weight to
observations whose contexts are close to a target context
$\CtxStar$.

For each observation, define the likelihood contribution
\[
L_i(\Param)
:=
p_{\rvObsi{i}\mid\rvParam,\rvCtxi{i}}
\left(
\Obsi{i}\mid\Param,\Ctxi{i}
\right).
\]
For a fixed prescribed localization rate, let
$\Bandwidth(\CtxStar)$ denote the bandwidth selected
at $\CtxStar$, and let $\AlgWeighti{i}\geq0$,
$i=1,\ldots,n$, be the corresponding weights satisfying
$\sum_{i=1}^n\AlgWeighti{i}=1$.

The localized posterior is defined by
\begin{equation}\label{eq:localized_posterior}
\LocalPost(\Param\mid\Dataset_n)
\propto
p_{\rvParam}(\Param)
\exp\!\left\{
\TauLoc{\Bandwidth(\CtxStar)}{\CtxStar}{\Ctxn}
\sum_{i=1}^{n}
\AlgWeighti{i}\log L_i(\Param)
\right\}.
\end{equation}

The weights determine the relative contribution of each
observation to calibration at $\CtxStar$. Since they sum
to one, the weighted log-likelihood does not have the
scale of the full-data log-likelihood. We therefore introduce
the positive scaling factor
$\TauLoc{\Bandwidth(\CtxStar)}{\CtxStar}{\Ctxn}$, based on the
effective sample size of the localization weights.
The weights and scaling factor are defined below.

\subsection{Localization weights}
\label{sec:localization_weights}

We measure proximity between contexts using a Gaussian
kernel with a candidate bandwidth $\Bandwidth>0$.
Let
\[
\CtxScaleMat
=
\operatorname{diag}
\left(
\ScaleCtxi{0},\ldots,\ScaleCtxi{\Ctxdim-1}
\right),
\]
where $\ScaleCtxi{k}>0$ is a characteristic scale of
context coordinate $k$.
The bandwidth covariance is
$\BandwidthMat_\Bandwidth=\Bandwidth^2\CtxScaleMat^2$.

The Gaussian localization kernel is
\begin{equation}\label{eq:gaussian_localization_weights}
\Kernelh(\CtxStar,\Ctxi{i})
\propto
\exp\!\left[
-\tfrac12(\CtxStar-\Ctxi{i})^\top
\BandwidthMat_\Bandwidth^{-1}
(\CtxStar-\Ctxi{i})
\right].
\end{equation}

When the recorded contexts are uncertain, we account for
their Gaussian errors through convolution of the localization
kernel \cite{carroll2006measurement}. This broadens the
kernel by replacing $\BandwidthMat_\Bandwidth$ with
$\BandwidthMat_\Bandwidth+\rvCtxNoiseCov$.

For diagonal context-error covariance
\[
\rvCtxNoiseCov
=
\operatorname{diag}
\left(q_0^2,\ldots,q_{\Ctxdim-1}^2\right),
\]
the resulting kernel is
\begin{equation}\label{eq:gaussian_localization_weights_noise}
\Kernelh^{\mathrm{noise}}(\CtxStar,\Ctxi{i})
\propto
\exp\!\left[
-\tfrac12\sum_{k=0}^{\Ctxdim-1}
\frac{
\bigl(\ctxi{k}^{\star}-(\Ctxi{i})_k\bigr)^2
}{
\Bandwidth^2\ScaleCtxi{k}^2+q_k^2
}
\right].
\end{equation}
Context uncertainty therefore reduces the degree of
localization achievable for a given bandwidth.

When the context domain is bounded, a Gaussian kernel may
also exhibit boundary effects. For a rectangular domain
$\CtxSpace=\prod_{k=0}^{\Ctxdim-1}[a_k,b_k]$,
we use a first-order reflection
\cite{schuster1985incorporating,botev2010kernel}.

Let $\Kernelh^{\mathrm{base}}$ denote either the Gaussian
kernel $\Kernelh$ or its noise-inflated version
$\Kernelh^{\mathrm{noise}}$. The reflected kernel is
\begin{equation}\label{eq:reflected_kernel}
\Kernelh^{\mathrm{ref}}(\CtxStar,\Ctxi{i})
=
\Kernelh^{\mathrm{base}}(\CtxStar,\Ctxi{i})
+
\sum_{k=0}^{\Ctxdim-1}
\left[
\Kernelh^{\mathrm{base}}(\Ctx_{a,k}^{\star},\Ctxi{i})
+
\Kernelh^{\mathrm{base}}(\Ctx_{b,k}^{\star},\Ctxi{i})
\right],
\end{equation}
where $\Ctx_{a,k}^{\star}$ and $\Ctx_{b,k}^{\star}$
are obtained by replacing coordinate $k$ of $\CtxStar$
with $2a_k-\ctxi{k}^{\star}$ and
$2b_k-\ctxi{k}^{\star}$, respectively.
This first-order correction includes reflections across
the domain faces. Its effect is illustrated in
\cref{app:kernel_illustrations}.

Let $\Kernelh^{\mathrm{sel}}$ denote the selected kernel,
with or without noise inflation and reflection.
For each candidate bandwidth $\Bandwidth$ and target
context $\CtxStar$, the normalized weights are
\begin{equation}\label{eq:normalized_localization_weights}
w_i^{(0)}(\Bandwidth(\CtxStar);\CtxStar,\Ctxn)
=
\frac{
\Kernelh^{\mathrm{sel}}(\CtxStar,\Ctxi{i})
}{
\sum_{j=1}^{n}
\Kernelh^{\mathrm{sel}}(\CtxStar,\Ctxi{j})
}.
\end{equation}

If the recorded contexts are unevenly distributed,
over-represented regions may dominate the weighted
likelihood. We therefore consider an optional correction
based on the context sampling density:
\begin{equation}\label{eq:flattened_localization_weights}
\widetilde w_i(\Bandwidth(\CtxStar);\CtxStar,\Ctxn)
=
\frac{
w_i^{(0)}(\Bandwidth(\CtxStar);\CtxStar,\Ctxn)/
\widehat p_n(\Ctxi{i})^\gamma
}{
\sum_{j=1}^{n}
w_j^{(0)}(\Bandwidth(\CtxStar);\CtxStar,\Ctxn)/
\widehat p_n(\Ctxi{j})^\gamma
},
\qquad 0\leq\gamma\leq1,
\end{equation}
where $\widehat p_n$ is an estimate of the context sampling
density, or the known density when available, assumed
positive at the recorded contexts.

Hereafter, $\Weighti{i}$ denotes the final normalized
weight for a candidate bandwidth: $w_i^{(0)}$ without
density correction and $\widetilde w_i$ when the
correction is applied.

Once the bandwidth $\Bandwidth(\CtxStar)$ has been
selected, the weights used in the localized posterior are
\begin{equation}\label{eq:final_localization_weights}
\AlgWeighti{i}
:=\Weighti{i},
\qquad
\sum_{i=1}^{n}\AlgWeighti{i}=1.
\end{equation}

\subsection{Posterior scaling}
\label{sec:posterior_scaling}

The scaling factor in \cref{eq:localized_posterior}
controls the contribution of the weighted likelihood
relative to the prior. For a candidate bandwidth, we
define the Kish effective sample size
\cite{kish1992weighting} by
\begin{equation}\label{eq:kish-ess}
\ESSLoc{\Bandwidth(\CtxStar)}{\CtxStar}{\Ctxn}
=
\frac{1}{
\sum_{i=1}^{n}
\left(\Weighti{i}\right)^2
}.
\end{equation}

The scaling factor used in the localized posterior is
\begin{equation}\label{eq:posterior_scaling}
\TauLoc{\Bandwidth(\CtxStar)}{\CtxStar}{\Ctxn}
=
\ESSLoc{\Bandwidth(\CtxStar)}{\CtxStar}{\Ctxn}.
\end{equation}

For uniform weights $\AlgWeighti{i}=1/n$, the effective
sample size is $n$, and the scaled weighted
log-likelihood coincides with the full-data
log-likelihood. If all the weight is assigned to
a single observation, the effective sample size
is one, and the localized likelihood reduces to
that observation's contribution.

\subsection{Bandwidth selection principle}
\label{sec:bandwidth_selection}

The bandwidth $\Bandwidth$ controls the strength of
localization. Although the context scales are incorporated
through $\CtxScaleMat$, the bandwidth itself does not
directly indicate how many observations contribute
effectively at a target context.

We therefore define the localization rate, expressed as
a percentage, by
\begin{equation}\label{eq:localizationrate}
\LocRate{\Bandwidth(\CtxStar)}{\CtxStar}{\Ctxn}
:=
\left(
1-
\frac{
\TauLoc{\Bandwidth(\CtxStar)}{\CtxStar}{\Ctxn}
}{n}
\right)\times100.
\end{equation}

Since
$1\leq\ESSLoc{\Bandwidth(\CtxStar)}{\CtxStar}{\Ctxn}\leq n$,
the localization rate satisfies
\[
0\leq
\LocRate{\Bandwidth(\CtxStar)}{\CtxStar}{\Ctxn}
\leq100(1-1/n).
\]
A rate of zero corresponds to uniform weights, while
larger values indicate stronger concentration of
the weights on a smaller effective number of observations.

For a prescribed feasible localization rate, we determine
a bandwidth separately for each target context.
Specifically, $\Bandwidth(\CtxStar)$ is chosen to satisfy
\begin{equation}\label{eq:target_bandwidth}
\LocRate{\Bandwidth(\CtxStar)}{\CtxStar}{\Ctxn}
=
\LocRateTarget.
\end{equation}
The bandwidth is computed numerically using Brent's method,
provided that a root is bracketed. The prescribed
localization rate is selected using predictive performance
metrics. The computation of the localization weights and
scaling factors is summarized in
\cref{alg:localization_precomputation}.

\subsection{Conditional flow and variational objective}
\label{sec:conditional_flow}

The preceding construction defines a family of localized
posteriors indexed by the target context. Approximating
each distribution separately would require solving
a new inference problem for every target context.
We instead use a conditional normalizing flow to
amortize inference over the context domain.

Let $z\sim\BaseLatent$ be a base random variable.
A conditional normalizing flow transforms $z$ through
an invertible map $\FlowMapVI(z;\CtxStar)$, parameterized
by $\VarParam$. The resulting conditional density is
\begin{equation}\label{eq:conditional_flow}
\qvar(\Param\mid \CtxStar)
=
\BaseLatent\bigl(\FlowMapVIInv(\Param;\CtxStar)\bigr)
\left|
\det D_{\Param}
\FlowMapVIInv(\Param;\CtxStar)
\right|.
\end{equation}
The conditioner networks receive $\CtxStar$ as an
additional input
\cite{papamakarios2017masked,kingma2016improved,dinh2014nice,dinh2016density}.

For each target context, the flow is trained to approximate
the localized posterior by minimizing the negative
evidence lower bound:
\begin{equation}\label{eq:localized_nelbo}
\AlgLocNELBOof{\VarParam}{\CtxStar}
=
\E_{\qvarof{\cdot\mid\CtxStar}}
\Bigg[
\log\qvarof{\rvParam\mid\CtxStar}
-
\log p_{\rvParam}(\rvParam)
-
\TauLoc{\Bandwidth(\CtxStar)}{\CtxStar}{\Ctxn}
\sum_{i=1}^{n}
\AlgWeighti{i}\log L_i(\rvParam)
\Bigg].
\end{equation}

Minimizing this objective is equivalent to minimizing
\[
\DKL{
\qvarof{\cdot\mid\CtxStar}
}{
\LocalPost(\cdot\mid\Dataset_n)
},
\]
since the two objectives differ by a constant independent
of $\VarParam$. The derivation is given in
\cref{app:elbo_procedures}.

To amortize the approximation, we consider a collection
of $n_{\mathrm{target}}$ target contexts $\Ctx_j^\star$,
each with its bandwidth
$\Bandwidth_j=\Bandwidth(\Ctx_j^\star)$.
We minimize the average negative ELBO,
\[
\frac{1}{n_{\mathrm{target}}}
\sum_{j=1}^{n_{\mathrm{target}}}
\AlgLocNELBOof{\VarParam}{\Ctx_j^\star}.
\]
For each target context, the bandwidth, localization
weights, and scaling factor are precomputed.

For a target mini-batch $B$, we draw $\nparticle$
parameter particles
$\Param_{jk}\sim\qvar(\cdot\mid\Ctx_j^\star)$.
For each target $j$, let $w_{ji}$ denote the final
localization weight of observation $i$, and define
\[
\Bandwidth_j=\Bandwidth(\Ctx_j^\star),
\qquad
w_{ji}
=
\Weighti{i},
\qquad
\ScalingFactor_j
=
\TauLoc{\Bandwidth_j}{\Ctx_j^\star}{\Ctxn}.
\]
We draw $\bobs$ observation indices independently from
\[
I_{j\ell}
\sim
\operatorname{Cat}(w_{j1},\ldots,w_{jn}).
\]
The stochastic negative ELBO is
\begin{align}\label{eq:mcmc_localized_elbo_loss}
\AlgLocNELBOMCof{\VarParam}{B}
=
\frac1{|B|\nparticle}
\sum_{j\in B}
\sum_{k=1}^{\nparticle}
\Bigl[
&\log\qvar(\Param_{jk}\mid\Ctx_j^\star)
-\log p_{\rvParam}(\Param_{jk})
\nonumber\\
&-\frac{\ScalingFactor_j}{\bobs}
\sum_{\ell=1}^{\bobs}
\log L_{I_{j\ell}}(\Param_{jk})
\Bigr].
\end{align}

Sampling observation indices according to the
localization weights gives an unbiased Monte Carlo
estimate of the weighted log-likelihood sum.
Parameter particles are generated through the flow
transformation, allowing reparameterization-based
differentiation.

If each epoch visits all $n_{\mathrm{target}}$ target
contexts, the number of model evaluations scales as
$\nepoch n_{\mathrm{target}}\nparticle\bobs$.
The precomputation and training algorithms are given
in \cref{app:elbo_procedures}.

\subsection{Aggregation under a prescribed context measure}
\label{sec:context_aggregation}

The conditional flow provides a parameter distribution
for each target context, with localization determined by
its corresponding bandwidth $\Bandwidth(\CtxStar)$.
To obtain a parameter distribution relevant to the
contexts specified by $\PCtx$, we integrate the
localized posteriors over this measure:
\begin{equation}\label{eq:context_marginalization}
p_{\mathrm{rob}}(\Param\mid\Dataset_n)
=
\int_{\CtxSpace}
\LocalPost(\Param\mid\Dataset_n)
\,\dd\PCtx(\CtxStar).
\end{equation}
The resulting distribution is a mixture of localized
posteriors weighted by the prescribed context measure.
It is approximated by drawing target contexts from
$\PCtx$ and parameter samples from the corresponding
conditional flow.

\section{Reference methods and predictive evaluation}
\label{sec:reference_methods_and_evaluation}

We compare the predictive performance of localized and global calibration across contexts. We consider both point predictions and ensemble
predictions, followed by aggregate error metrics over test contexts.

\subsection{Reference methods}
\label{sec:reference_methods}

We consider two reference methods based on the global posterior,
which uses all observations without localization. The first is
deterministic calibration using the maximum a posteriori (MAP)
estimate of the global posterior. The second approximates the
global posterior using variational inference with a normalizing
flow.

For posterior-based inference, we consider two prediction
approaches. Point predictions are obtained by propagating a
parameter estimate, either the posterior mode (MAP) or the
componentwise posterior median, through $\ModelObs$.
Ensemble predictions are obtained by propagating posterior
parameter samples through the same model, allowing the
resulting predictive uncertainty to be quantified.

\subsection{Point and ensemble errors}
\label{sec:prediction_errors}

All experiments are based on twin experiments, for which
the noise-free physical response is available for evaluation:
$\RefObsi{i}=\ModelObsPhys(\CtxTruei{i})$.
However, the true context $\CtxTruei{i}$ is not available
to the calibration or prediction procedure.
Predictions are therefore obtained by evaluating the
numerical model at the recorded noisy context $\Ctxi{i}$.

For a point estimate $\widehat{\Param}$, the prediction is
$\ModelObsCall{\widehat{\Param}}{\Ctxi{i}}$.
For ensemble predictions, parameter samples are propagated
through the same model at $\Ctxi{i}$.
Both are compared with the reference response
$\RefObsi{i}$ using the absolute and relative discrepancies
\begin{equation}\label{eq:evaluation_rmse}
d_{\mathrm{abs}}(a,b)=\frac{\|a-b\|_2}{\sqrt{\Obsdim}},
\qquad
d_{\mathrm{rel}}(a,b)=\frac{\|a-b\|_2}{\|b\|_2},
\quad \|b\|_2>0.
\end{equation}

For point predictions, the error is the chosen discrepancy
between the propagated parameter estimate and the reference
response. For ensemble predictions, let
$\Obsi{i}^{(m)}$, $m=1,\ldots,\npred$, denote the
predictive samples at recorded test context $\Ctxi{i}$.
We use the criterion
\begin{equation}\label{eq:ens_error}
e_i(d)=\frac1{\npred}\sum_{m=1}^{\npred}
d(\Obsi{i}^{(m)},\RefObsi{i})
-\frac1{2\npred^2}\sum_{m,m'=1}^{\npred}
d(\Obsi{i}^{(m)},\Obsi{i}^{(m')}).
\end{equation}
For $d=d_{\mathrm{abs}}$, this is the empirical energy
score, scaled by $1/\sqrt{\Obsdim}$, and reduces to the
empirical continuous ranked probability score (CRPS)
in one dimension
\cite{gneiting2007strictly,gneiting2005calibrated,mills2025exploring}.
\subsection{Aggregate metrics}
\label{sec:aggregate_metrics}

We evaluate predictive performance over $\ntest$ test
contexts. For each evaluation criterion, let
$E_1,\ldots,E_{\ntest}$ denote the corresponding
per-context errors. We report their mean $\overline e$,
empirical standard deviation $\ErrorStd$, and upper-tail
mean $\ErrorUpperTailMean{\alpha}$:
\begin{align}\label{eq:aggregate_metrics}
 \overline e&=\frac1{\ntest}\sum_i e_i,&
 \ErrorStd&=\left[\frac1{\ntest-1}\sum_i(e_i-\overline e)^2\right]^{1/2},\nonumber\\
 \ErrorUpperTailMean{\alpha}&=\frac1{|A_\alpha|}\sum_{i\in A_\alpha}e_i,&
 A_\alpha&=\{i:e_i\geq Q_\alpha\}.
\end{align}
Here $Q_\alpha$ is the empirical $\alpha$-quantile
of the per-context errors. We use $\alpha=0.9$,
so $\ErrorUpperTailMean{0.9}$ measures the average
upper tail error.

\section{Branin--Hoo twin experiment}
\label{sec:branin}

We consider a numerical model based on the Branin--Hoo function,
a classical low-dimensional benchmark with a nonconvex structure
and several separated basins of attraction.

\subsection{Model, reference response, and data}
\label{sec:branin_design}

\begin{figure}[!htbp]
    \centering
\includegraphics[width=370.38104000pt,height=166.30420000pt]{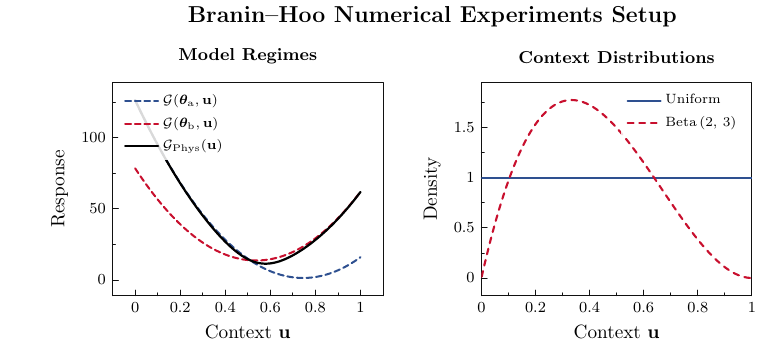}
\caption{Branin--Hoo experiment. Left: synthetic physical
response and the two regime-specific model
responses. Right: uniform and $\BetaDist{2}{3}$
context distributions.}
\label{fig:branin_setup}
\end{figure}

For $\param,\ctx\in[0,1]$, the numerical model is
$\ModelObsCall{\param}{\ctx}=B(15\param-5,15\ctx)$, where
\begin{equation}\label{eq:branin_model}
 B(x_1,x_2)=\left(x_2-\frac{5.1x_1^2}{4\pi^2}+\frac{5x_1}{\pi}-6\right)^2
 +10\left(1-\frac1{8\pi}\right)\cos x_1+10.
\end{equation}
The synthetic physical response interpolates between two parameter regimes:
\begin{align}\label{eq:branin_physical_model}
 \ModelObsPhys(\ctx)&=\ModelObsCall{\param_a}{\ctx}
 +s(\ctx)\bigl[\ModelObsCall{\param_b}{\ctx}
 -\ModelObsCall{\param_a}{\ctx}\bigr],\nonumber\\
 s(\ctx)&=\frac1{1+\exp[-(\ctx-\ctx_{\mathrm{Trans}})/\TransWidth]}.
\end{align}
The parameters are $\param_a=0.155$, $\param_b=0.255$,
$\ctx_{\mathrm{Trans}}=0.55$, and $\TransWidth=0.08$.
The transition prevents a single calibration parameter from
representing both reference regimes.
\Cref{fig:branin_setup} shows the model responses and context distributions.

Calibration contexts are sampled from either $\Unif{[0,1]}$
or $\BetaDist{2}{3}$, the latter concentrating observations
toward smaller context values. Both distributions are also
used for evaluation, giving four combinations of calibration
and test distributions. The observation-noise variance is
$5^2$, and the context-noise variance is $0.05^2$.
The calibration, test, and training-target sample sizes
are 64, 16, and 100, respectively. Training target contexts
are sampled uniformly, and each predictive ensemble contains
64 parameter samples. Predictions use the recorded noisy
contexts, as described in \cref{sec:prediction_errors}.

\subsection{Inference settings}
\label{sec:branin_inference}

The prior is uniform on $[0,1]$. The deterministic MAP
optimization starts at $0.5$. All methods use Adam with
an initial learning rate of $10^{-2}$ and 1500 iterations.
The global and conditional normalizing flows use three
spline coupling layers, eight spline bins, and a spline
bound of 1. The conditional network has two hidden layers
of width 32. Variational training uses 16 particles and
a target-context batch size of 64.

\subsection{Sensitivity to localization rate}
\label{sec:branin_rate}

\begin{figure}[!htbp]
  \centering
\includegraphics[width=370.39589000pt,height=292.68039000pt]{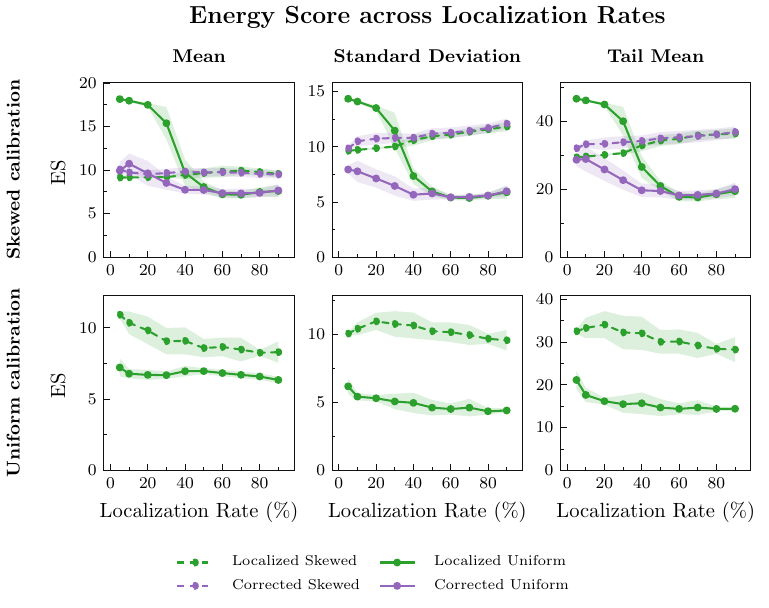}
  \caption{
  Evolution of the energy-score metrics as a function of the localization rate. The first row corresponds to calibration with skewed context data, while the second row corresponds to calibration with uniformly distributed context data. Columns report the mean, standard deviation, and tail mean over the evaluation errors. Shaded regions indicate variability across the reported random-seed runs.
  }
  \label{fig:lines_metrics_per_rate}
\end{figure}

\Cref{fig:lines_metrics_per_rate} reports the three
absolute ensemble metrics across the localization-rate
sweep, with one-standard-deviation envelopes over seeds.
Under skewed calibration sampling, stronger localization
decreases errors on uniformly distributed test contexts
but can increase them on skewed test contexts.
The skewed test distribution emphasizes the lower-context
regime, where calibration dominated by that regime is
less penalized. In contrast, uniform evaluation also
covers the transition and the second regime.
Under uniform calibration sampling, the reported metrics
decrease with localization for both test distributions.

\subsection{Aggregate ensemble performance}
\label{sec:branin_aggregates}

\Cref{tab:ensemble_metrics} reports the mean, standard
deviation across test contexts, and upper-tail mean
of the absolute energy score.
For uniform calibration and uniform testing, the mean
decreases from $43.537\pm19.324$ for the global posterior
to $6.679\pm0.265$ and $6.699\pm0.222$ at localization
rates of $30\%$ and $70\%$, respectively.
For skewed calibration and uniform testing, the corresponding
values are $43.846\pm17.640$, $15.382\pm1.857$, and
$7.169\pm0.403$. The localized methods also reduce the
reported upper-tail errors relative to the global posterior.

The effect of density correction depends on the localization
rate and test distribution. Under skewed calibration and
uniform testing, it improves the mean at $30\%$ from
$15.382$ to $8.525$, but the corrected mean at $70\%$
is $7.334$, compared with $7.169$ without correction.
Under skewed testing, the correction increases the mean
at $30\%$ from $9.201$ to $9.665$ and decreases it at
$70\%$ from $9.936$ to $9.691$.

\begin{table}[!htbp]\centering\small
\begin{tabular}{lccc}\toprule
Method & $\ErrorMean$ & $\ErrorStd$ & $\ErrorUpperTailMean{0.9}$ \\
\midrule\multicolumn{4}{l}{Uniform calibration; uniform test contexts} \\
Global & \num{43.537} $\pm$ \num{19.324} & \num{22.919} $\pm$ \num{8.109} & \num{82.614} $\pm$ \num{30.431} \\
Localized $30\%$ & \num{6.679} $\pm$ \num{0.265} & \num{5.071} $\pm$ \num{0.567} & \num{15.500} $\pm$ \num{2.037} \\
Localized $70\%$ & \num{6.699} $\pm$ \num{0.222} & \num{4.614} $\pm$ \num{0.647} & \num{14.695} $\pm$ \num{1.703} \\
\midrule\multicolumn{4}{l}{Uniform calibration; skewed test contexts} \\
Global & \num{32.138} $\pm$ \num{13.502} & \num{22.571} $\pm$ \num{8.638} & \num{70.060} $\pm$ \num{27.025} \\
Localized $30\%$ & \num{9.063} $\pm$ \num{0.929} & \num{10.770} $\pm$ \num{0.942} & \num{32.279} $\pm$ \num{3.927} \\
Localized $70\%$ & \num{8.471} $\pm$ \num{0.837} & \num{9.956} $\pm$ \num{0.732} & \num{29.227} $\pm$ \num{2.937} \\
\midrule\multicolumn{4}{l}{Skewed calibration; uniform test contexts} \\
Global & \num{43.846} $\pm$ \num{17.640} & \num{23.770} $\pm$ \num{6.462} & \num{84.244} $\pm$ \num{25.942} \\
Localized $30\%$ & \num{15.382} $\pm$ \num{1.857} & \num{11.468} $\pm$ \num{1.602} & \num{39.957} $\pm$ \num{4.217} \\
Localized corrected $30\%$ & \num{8.525} $\pm$ \num{0.772} & \num{6.453} $\pm$ \num{0.869} & \num{22.649} $\pm$ \num{2.841} \\
Localized $70\%$ & \num{7.169} $\pm$ \num{0.403} & \num{5.361} $\pm$ \num{0.222} & \num{17.545} $\pm$ \num{1.156} \\
Localized corrected $70\%$ & \num{7.334} $\pm$ \num{0.551} & \num{5.462} $\pm$ \num{0.237} & \num{18.305} $\pm$ \num{1.351} \\
\midrule\multicolumn{4}{l}{Skewed calibration; skewed test contexts} \\
Global & \num{30.637} $\pm$ \num{14.464} & \num{21.951} $\pm$ \num{8.659} & \num{68.719} $\pm$ \num{27.332} \\
Localized $30\%$ & \num{9.201} $\pm$ \num{0.084} & \num{10.024} $\pm$ \num{0.143} & \num{30.591} $\pm$ \num{0.503} \\
Localized corrected $30\%$ & \num{9.665} $\pm$ \num{0.480} & \num{10.799} $\pm$ \num{0.454} & \num{33.792} $\pm$ \num{1.795} \\
Localized $70\%$ & \num{9.936} $\pm$ \num{0.487} & \num{11.369} $\pm$ \num{0.324} & \num{35.770} $\pm$ \num{1.396} \\
Localized corrected $70\%$ & \num{9.691} $\pm$ \num{0.474} & \num{11.451} $\pm$ \num{0.476} & \num{35.607} $\pm$ \num{1.852} \\
\bottomrule\end{tabular}
\caption{Absolute ensemble energy-score aggregates. The columns give the mean, standard deviation across test contexts, and upper-tail mean. Each cell is the mean $\pm$ standard deviation over the reported seed runs.}
\label{tab:ensemble_metrics}\end{table}

\subsection{Predictions and marginalized parameter distributions}
\label{sec:branin_predictions}

\begin{figure}[!htbp]
    \centering
\includegraphics[width=370.40462000pt,height=334.40286000pt]{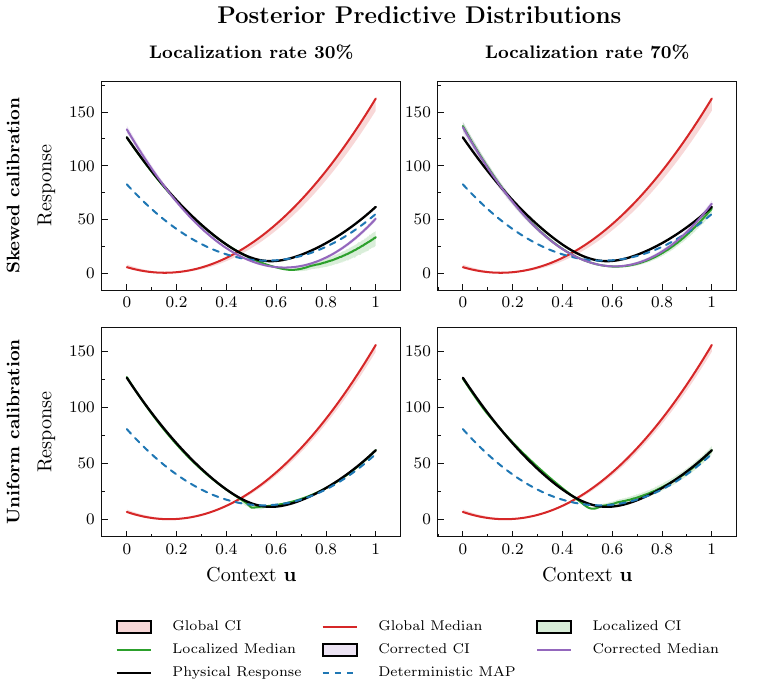}
    \caption{Posterior predictive distributions for the Branin--Hoo benchmark at localization rates $30\%$ and $70\%$. Rows correspond to calibration with skewed and uniform context distributions, while columns compare the two localization rates. Each panel shows the predictive $90\%$ credible intervals, posterior medians, physical response, and deterministic MAP prediction.}
    \label{fig:branin_predictive_posteriors_30_70}
\end{figure}

The posterior predictive medians and $90\%$ credible
intervals in \cref{fig:branin_predictive_posteriors_30_70}
show that localized inference follows the change in
physical response more closely than a single global
calibration.

\begin{figure}[!htbp]
    \centering
\includegraphics[width=370.39906000pt,height=301.05191000pt]{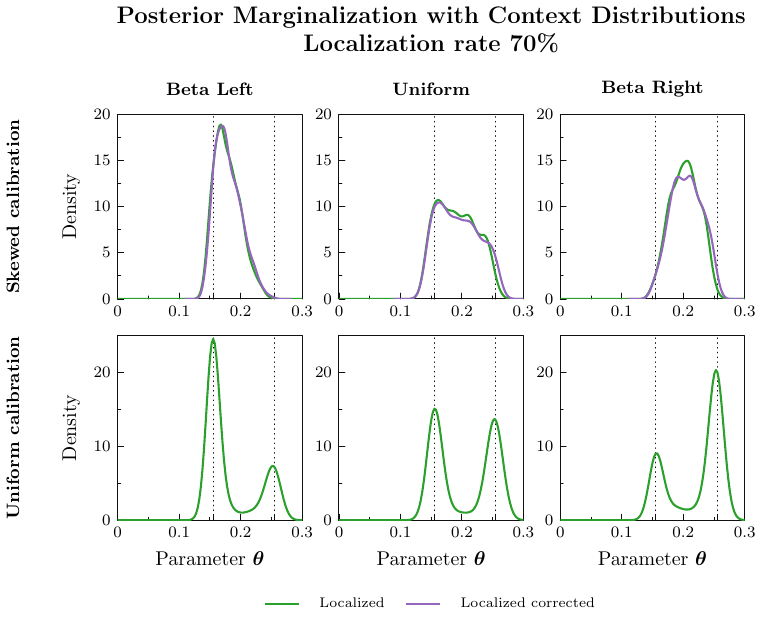}
    \caption{Marginalized posterior distributions for localization rate $70\%$, obtained by integrating the localized posterior over different context distributions. The first row corresponds to calibration with skewed context data, while the second row corresponds to calibration with uniform context data. Columns show marginalization with respect to context distributions concentrated toward smaller values, uniform contexts, and distributions concentrated toward larger values. Vertical dotted lines indicate the reference parameter values associated with the two regimes.}
    \label{fig:marginal_posteriors_70}
\end{figure}

\Cref{fig:marginal_posteriors_70} compares the marginalized
parameter distributions under $\BetaDist{2}{3}$,
the uniform measure, and $\BetaDist{3}{2}$.
These measures emphasize smaller contexts, the full domain,
and larger contexts, respectively.
The marginalized distribution changes with the prescribed
context measure. Under skewed calibration, however, the
sampling imbalance continues to influence its structure.
Uniform calibration gives a clearer separation between
the contributions of the two parameter regimes.
The corresponding marginalized distributions at $30\%$
localization are presented in \cref{app:branin_results}.

\FloatBarrier

\section{Ocean turbulence-closure twin experiment}
\label{sec:ocean_turbulence_experiment}

We apply amortized localized posterior calibration to a single-column ocean model implemented in Jax by \cite{tunax_2024} from \cite{perrot2024energetically}.
The $k$--$\DissipationRate$ closure \cite{launder1974application}
is implemented in the generic length-scale framework
\cite{umlauf2003generic}.

\subsection{Single-column model, transport, and surface forcing}
\label{sec:single_column_transport}

The vertical coordinate $\Depth\in[-\DepthMax,0]$ increases upward.
The state is $X(t,\Depth)=(\ZonalVel,\Temp,\Sal)$, consisting of
zonal velocity, temperature, and salinity. Coriolis frequency,
meridional forcing, and initial meridional velocity are zero,
so the meridional velocity remains zero.
Density follows the linear equation of state
\begin{equation}\label{eq:scm_linear_eos}
\Density=\RefDensity\bigl[
1-\alpha(\Temp-\Temp_{\mathrm{ref}})
+\beta(\Sal-\Sal_{\mathrm{ref}})
\bigr].
\end{equation}
Vertical turbulent diffusivity $\DiffVert$ and viscosity
$\ViscVert$ govern transport:
\begin{equation}\label{eq:scm_transport_temperature}
\partial_t\Temp=\partial_{\Depth}(\DiffVert\partial_{\Depth}\Temp),
\quad
\partial_t\Sal=\partial_{\Depth}(\DiffVert\partial_{\Depth}\Sal),
\quad
\partial_t\ZonalVel=\partial_{\Depth}(\ViscVert\partial_{\Depth}\ZonalVel).
\end{equation}
At the surface,
\begin{equation}\label{eq:scm_surface_boundary_conditions}
\DiffVert\partial_{\Depth}\Temp=\HeatForcing,\qquad
\DiffVert\partial_{\Depth}\Sal=0,\qquad
\ViscVert\partial_{\Depth}\ZonalVel=\WindForcing.
\end{equation}
Positive $\HeatForcing$ represents heat entering the ocean,
whereas negative $\HeatForcing$ represents surface cooling.
Positive $\WindForcing$ accelerates the positive zonal velocity.
Bottom tracer fluxes and stress are zero, with
$\partial_{\Depth}\Temp=\partial_{\Depth}\Sal
=\partial_{\Depth}\ZonalVel=0$
at $\Depth=-\DepthMax$.
The context is $\Ctx=(\WindForcing,\HeatForcing)^\top$.
Temperature and zonal velocity are the observed variables.

\subsection{Turbulence closure and calibrated parameters}
\label{sec:turbulence_closure}

The closure evolves turbulent kinetic energy $k$ and its
dissipation rate $\DissipationRate$:
\begin{align}\label{eq:scm_tke_equation}
\partial_tk
&=
\partial_{\Depth}\!\left(
\frac{\ViscVert}{\sigma_k}\partial_{\Depth}k
\right)
+\ShearProduction+\BuoyancyProduction-\DissipationRate,\\
\partial_t\DissipationRate
&=
\partial_{\Depth}\!\left(
\frac{\ViscVert}{\sigma_\DissipationRate}
\partial_{\Depth}\DissipationRate
\right)
+\frac\DissipationRate k
\left(
\ParamCeps\ShearProduction
+c_{\DissipationRate3}\BuoyancyProduction
-c_{\DissipationRate2}\DissipationRate
\right).
\label{eq:scm_epsilon_equation}
\end{align}
The production terms and mixing parameters are
\begin{align}
\ShearProduction
&=\ViscVert(\partial_{\Depth}\ZonalVel)^2,
&
\BuoyancyProduction
&=-\DiffVert N^2,
&
N^2
&=-\frac g{\RefDensity}\partial_{\Depth}\Density,
\nonumber\\
\ViscVert
&=c_\mu\frac{k^2}{\DissipationRate},
&
\DiffVert
&=c_\mu'\frac{k^2}{\DissipationRate}.
\label{eq:scm_mixing_parameters}
\end{align}
The stability functions $c_\mu$ and $c_\mu'$ depend algebraically
on $\alpha_M=(k^2/\DissipationRate^2)
(\partial_{\Depth}\ZonalVel)^2$
and $\alpha_N=(k^2/\DissipationRate^2)N^2$.
Their parameters are fixed. Stable stratification gives
$\BuoyancyProduction<0$, whereas unstable stratification gives
$\BuoyancyProduction>0$; $c_{\DissipationRate3}$ takes its
prescribed stable or unstable value.

The calibrated parameters are
$\Param=(\ParamCmu,\ParamCeps)$.
The parameter $\ParamCeps$ weights shear production in
\cref{eq:scm_epsilon_equation}, while $\ParamCmu$ relates
the turbulent length scale $\ell$ to $k$ and $\DissipationRate$:
\begin{equation}\label{eq:scm_cmu0_relation}
\DissipationRate
=
\ParamCmu^3\frac{k^{3/2}}{\ell},
\qquad
\ParamCmu\in[\num{0.3},\num{0.8}],
\quad
\ParamCeps\in[\num{0.1},\num{1.9}].
\end{equation}
The parameter bounds were selected around the default values
while accounting for numerical stability.
The physical and closure settings are provided in
\cref{app:ocean_calibration_settings}, and
\cref{tab:kepsilon_default_parameters} lists the default
closure values and fixed stability-function parameters.

\subsection{Forcing-dependent reference parameters}
\label{sec:ocean_reference_parameters}

\begin{figure}[!htbp]
    \centering
        \includegraphics[width=\textwidth]{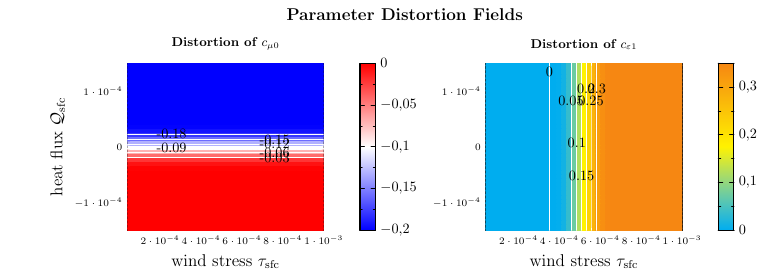}
    \caption{Illustration of the context-dependent parameter-space distortion used to represent model discrepancy in the twin experiment.}
    \label{fig:tunax_parameter_space_distortion}
\end{figure}

To construct the twin experiment, we generate the reference
response using the same numerical model with context-dependent
parameters:
\begin{equation}\label{eq:twin_true_parameters}
\ParamTrue(\Ctx)=\ParamRef+\Distorsion(\Ctx),
\qquad
\ModelObsPhys(\Ctx)=\ModelObsCall{\ParamTrue(\Ctx)}{\Ctx},
\qquad
\ParamRef=(\num{0.5477},\num{1.44}).
\end{equation}
Let $\overline{\WindForcing}$ and $\overline{\HeatForcing}$
denote the forcing coordinates scaled to $[0,1]$ using their
domain bounds. The imposed distortion is
\begin{equation}\label{eq:twin_parameter_distortion}
\Distorsion(\Ctx)
=
\begin{pmatrix}
-a_{\mu0}\mathcal S_\delta(\overline{\HeatForcing})\\
a_{\DissipationRate1}\mathcal S_\delta(\overline{\WindForcing})
\end{pmatrix},
\qquad
a_{\mu0}=\num{0.2},
\quad
a_{\DissipationRate1}=\num{0.35},
\quad
\delta=\num{0.15}.
\end{equation}
The smooth transition $\mathcal S_\delta$ is defined in
\cref{app:ocean_model_details}.
Thus, wind forcing changes $\ParamCeps$, while heat forcing
changes $\ParamCmu$, producing four forcing regimes.
\Cref{fig:tunax_parameter_space_distortion} illustrates
the imposed distortion.

\subsection{Data, uncertainty, and numerical configuration}
\label{sec:ocean_data_and_numerics}

Contexts are sampled uniformly over the forcing domain.
The calibration and test sets contain \num{1024} and \num{512}
contexts, respectively, and \num{512} target contexts are used
for amortized training.
Each response concatenates temperature and zonal velocity over
\num{100} depth cells and \num{73} stored times.
With $m=\num{100}\times\num{73}$, the observation-noise covariance is
\begin{equation}\label{eq:observation_noise_covariance}
\rvObsNoiseCov
=
\operatorname{diag}
(\sigma_\Temp^2I_m,\sigma_\ZonalVel^2I_m).
\end{equation}
Each observation-noise standard deviation is \SI{5}{\percent}
of the corresponding reference-signal standard deviation
in the calibration dataset.

Recorded contexts have independent Gaussian errors with covariance
\begin{equation}\label{eq:context_noise_covariance}
\rvCtxNoiseCov
=
\operatorname{diag}\left[
(\SI{4.8e-5}{\metre\squared\per\second\squared})^2,
(\SI{1.5e-5}{\kelvin\metre\per\second})^2
\right].
\end{equation}
These standard deviations are approximately \SI{5}{\percent}
of the respective forcing ranges.

The \SI{2000}{\metre} column uses \num{100} cells,
approximately uniform in the upper ocean and increasingly
thick below \SI{400}{\metre}, with stretching parameter
$\param_{\mathrm{grid}}=\num{6.5}$.
Each simulation lasts \SI{72}{\hour}, with a time step of
\SI{50}{\second} and output every \SI{3600}{\second}.
The \num{73} stored times include the initial state.
The initial profiles are
\begin{equation}\label{eq:scm_initial_profiles}
\ZonalVel(0,\Depth)=0,\qquad
\Temp(0,\Depth)
=
\SI{2}{\degreeCelsius}
+
(\SI{1e-3}{\kelvin\per\metre})\Depth,
\qquad
\Sal(0,\Depth)=\SI{32.6}{\psu}.
\end{equation}
Uniform salinity and zero salinity flux imply that
temperature controls the density gradient.
The surface-forcing domain is
\begin{align}\label{eq:scm_context_domain}
\WindForcing
&\in
[\SI{4e-5}{\metre\squared\per\second\squared},
 \SI{1e-3}{\metre\squared\per\second\squared}],
\nonumber\\
\HeatForcing
&\in
[\SI{-1.5e-4}{\kelvin\metre\per\second},
 \SI{1.5e-4}{\kelvin\metre\per\second}].
\end{align}

\subsection{Inference and predictive evaluation}
\label{sec:ocean_calibration_and_evaluation}

All methods use \num{45} epochs, an initial learning rate of
\num{1e-2}, and random seed \num{100}.
The flows use three layers, hidden dimensions
$(\num{64},\num{64},\num{64})$, eight spline bins,
a spline bound of \num{5}, and a gradient-clipping norm
of \num{0.5}.
Complete settings are provided in
\cref{app:ocean_calibration_settings}.

We evaluate temperature and zonal velocity separately using
the point and ensemble criteria defined in
\cref{sec:reference_methods_and_evaluation}.
For each variable, predictions and reference responses are
flattened over depth and time, giving an output dimension $m$.
Predictions use the recorded noisy contexts.
The reported aggregate metrics are computed over the
\num{512} test contexts.

The profile comparisons show posterior predictive medians
and \SI{90}{\percent} intervals, the twin reference,
and deterministic MAP predictions.
Normalized depth is computed using the mixed-layer
diagnostic described in \cref{app:ocean_model_details}.

\subsection{Aggregate errors and parameter distributions}
\label{sec:ocean_predictive_results}

\Cref{tab:predictive_metrics_scm} reports the point
and ensemble errors.
The deterministic MAP mean RMSE is
\SI{0.01833}{\kelvin} for temperature and
\SI{0.05737}{\metre\per\second} for zonal velocity.
Global-posterior point summaries give similar errors.
Localized-posterior median predictions reduce the means
to \SI{0.01160}{\kelvin} and
\SI{0.02826}{\metre\per\second} at
\SI{80}{\percent} localization, and to
\SI{0.01097}{\kelvin} and
\SI{0.02530}{\metre\per\second} at
\SI{98}{\percent} localization, respectively.
Localized MAP predictions are close to the median predictions.

The global mean ensemble scores are
\SI{0.01790}{\kelvin} and
\SI{0.05362}{\metre\per\second}.
At \SI{80}{\percent} localization, they decrease to
\SI{0.01134}{\kelvin} and
\SI{0.02725}{\metre\per\second};
at \SI{98}{\percent}, they decrease to
\SI{0.01044}{\kelvin} and
\SI{0.02305}{\metre\per\second}.
The standard deviations and upper-tail means also decrease.

\begin{table}[!htbp]\centering\small
\begin{tabular}{llccc}\toprule Method & Prediction & $\ErrorMean$ & $\ErrorStd$ & $\ErrorUpperTailMean{0.9}$ \\
\midrule\multicolumn{5}{l}{\textit{Temperature (K)}} \\
Deterministic MAP & MAP & \num{0.01833} & \num{0.01763} & \num{0.05766} \\
Global & Median & \num{0.01802} & \num{0.01874} & \num{0.06033} \\
Global & MAP & \num{0.01852} & \num{0.02098} & \num{0.06739} \\
Localized $80\%$ & Median & \num{0.01160} & \num{0.01140} & \num{0.03620} \\
Localized $80\%$ & MAP & \num{0.01161} & \num{0.01140} & \num{0.03622} \\
Localized $98\%$ & Median & \num{0.01097} & \num{0.01126} & \num{0.03506} \\
Localized $98\%$ & MAP & \num{0.01100} & \num{0.01135} & \num{0.03522} \\
Global & Ensemble & \num{0.01790} & \num{0.01868} & \num{0.06009} \\
Localized $80\%$ & Ensemble & \num{0.01134} & \num{0.01132} & \num{0.03577} \\
Localized $98\%$ & Ensemble & \num{0.01044} & \num{0.01107} & \num{0.03422} \\
\midrule\multicolumn{5}{l}{\textit{Zonal velocity ($\mathrm{m\,s^{-1}}$)}} \\
Deterministic MAP & MAP & \num{0.05737} & \num{0.02671} & \num{0.10678} \\
Global & Median & \num{0.05410} & \num{0.02826} & \num{0.10978} \\
Global & MAP & \num{0.05554} & \num{0.03193} & \num{0.12354} \\
Localized $80\%$ & Median & \num{0.02826} & \num{0.01656} & \num{0.06306} \\
Localized $80\%$ & MAP & \num{0.02826} & \num{0.01654} & \num{0.06299} \\
Localized $98\%$ & Median & \num{0.02530} & \num{0.01631} & \num{0.05969} \\
Localized $98\%$ & MAP & \num{0.02534} & \num{0.01620} & \num{0.05956} \\
Global & Ensemble & \num{0.05362} & \num{0.02820} & \num{0.10926} \\
Localized $80\%$ & Ensemble & \num{0.02725} & \num{0.01634} & \num{0.06162} \\
Localized $98\%$ & Ensemble & \num{0.02305} & \num{0.01572} & \num{0.05614} \\
\bottomrule\end{tabular}
\caption{Predictive error aggregates over 512 test contexts. Point predictions use absolute RMSE; ensembles use the absolute energy score.}
\label{tab:predictive_metrics_scm}\end{table}

\begin{figure}[!htbp]
    \centering
\includegraphics[width=307.30702000pt,height=344.42540000pt]{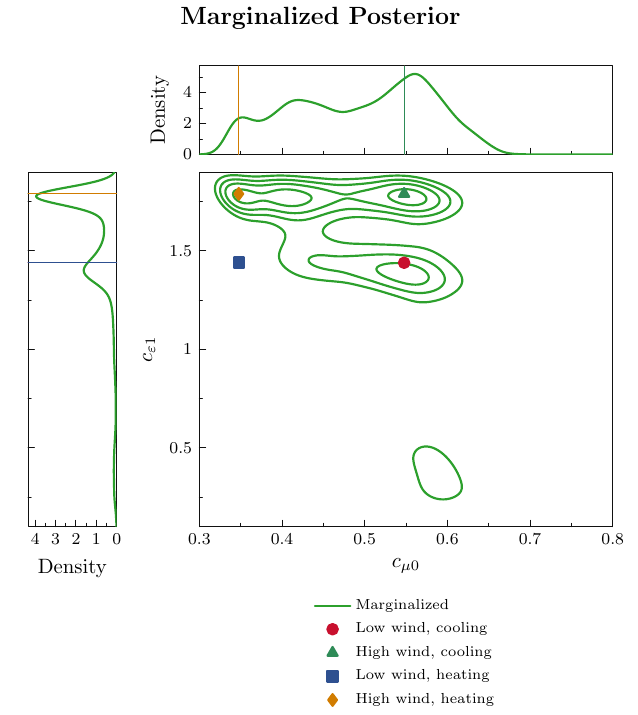}
    \caption{Marginalized posterior obtained by combining $32$ samples from the localized posterior at each of $\num{1000}$ contexts spanning the context domain. The contours are estimated using kernel density estimation. The localized posterior uses rate $98\%$.}
    \label{fig:marginalized_posterior}
\end{figure}

The marginalized localized posterior in
\cref{fig:marginalized_posterior} is estimated using
\num{32} parameter samples at each of \num{1000}
contexts and displayed through kernel density estimation.
Three modes correspond to weak-wind cooling,
strong-wind cooling, and strong-wind heating.
The weak-wind heating regime is less identifiable:
limited mixing under these stable, weakly forced conditions
reduces the response sensitivity to the closure parameters.

\subsection{Time-averaged profiles across forcing regimes}
\label{sec:ocean_time_averaged_profiles}

\begin{figure}[!htbp]
    \centering
\includegraphics[width=370.40497000pt,height=395.17836000pt]{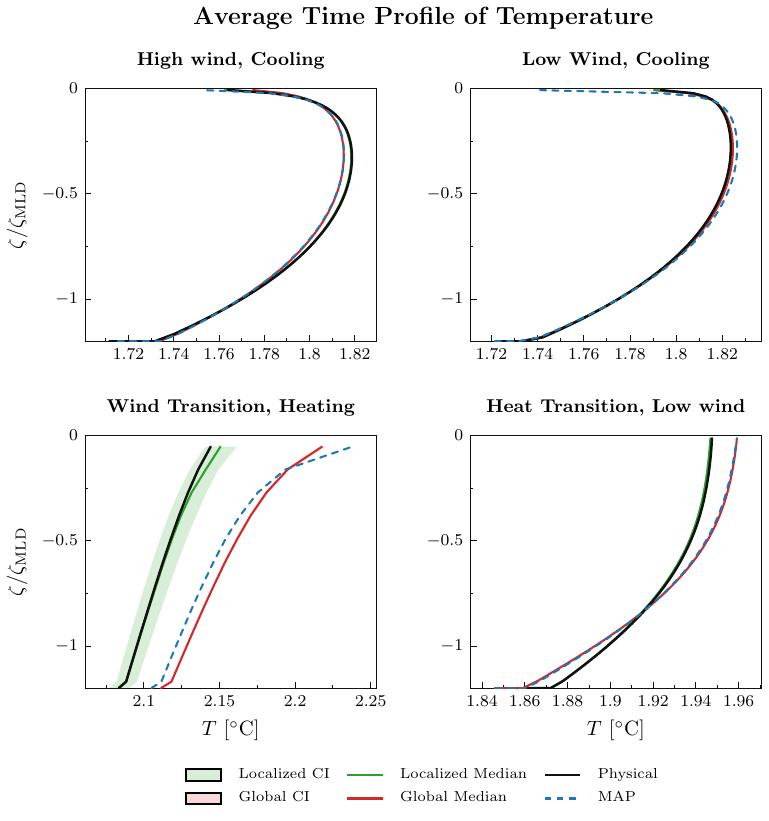}
    \caption{Time-averaged temperature profiles as functions of normalized depth $\Depth/\MLD$ for representative forcing regimes. The black curves denote the reference twin-simulation profiles. The green and red curves represent the posterior medians obtained from the localized and global posteriors, respectively, with shaded regions indicating the corresponding $90\%$ credible intervals. The blue dashed curves represent the predictions based on the maximum a posteriori estimates. The displayed depth range is restricted to the upper mixed-layer region, $\Depth/\MLD \in [-1.2,0]$. Localized curves use $98\%$ localization.}
    \label{fig:average_time_profile_temperature}
\end{figure}

\begin{figure}[!htbp]
    \centering
\includegraphics[width=370.40830000pt,height=395.17836000pt]{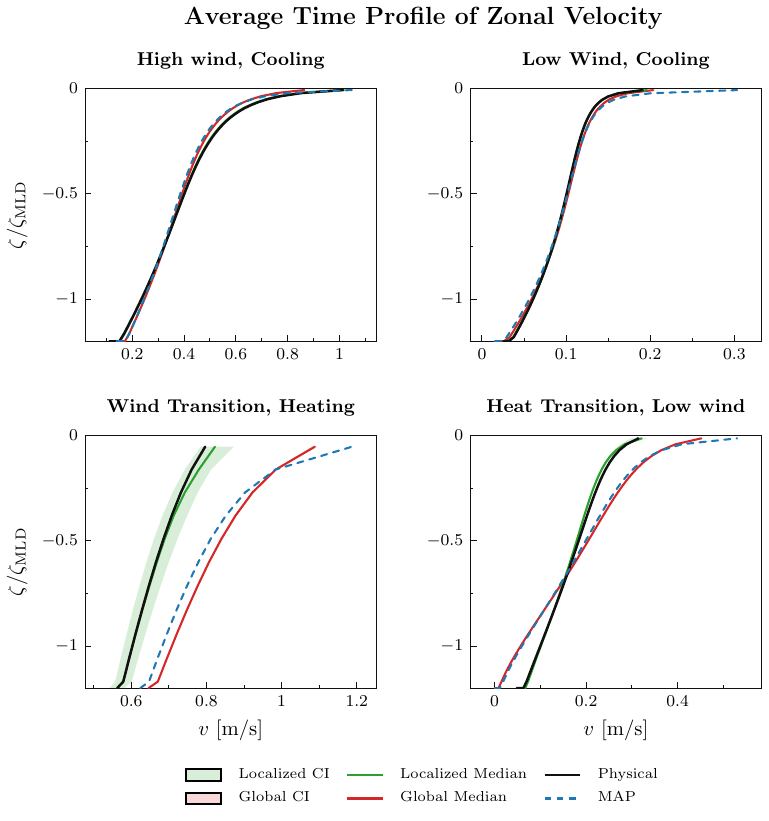}
    \caption{Time-averaged zonal-velocity profiles as functions of normalized depth $\Depth/\MLD$ for representative forcing regimes. The black curves denote the reference twin-simulation profiles. The green and red curves represent the posterior medians obtained from the localized and global posteriors, respectively, with shaded regions indicating the corresponding $90\%$ credible intervals. The blue dashed curves represent the predictions based on the maximum a posteriori estimates. The displayed depth range is restricted to the upper mixed-layer region, $\Depth/\MLD \in [-1.2,0]$. Localized curves use $98\%$ localization.}
    \label{fig:average_time_profile_zonal_velocity}
\end{figure}

\Cref{fig:average_time_profile_temperature,fig:average_time_profile_zonal_velocity}
show time-averaged profiles at \SI{98}{\percent}
localization across the selected forcing regimes.
Global calibration remains farther from the twin profiles,
whereas localized predictions adapt more closely to the
forcing context. The localized predictive intervals widen
in transitions between regimes, where nearby contexts
support different adequate parameter values.
The final-time comparisons after \SI{72}{\hour},
presented in \cref{app:final_profiles}, show the same
behavior for both variables.


\FloatBarrier

\section{Discussion}\label{sec:discussion}

In the Branin--Hoo experiment, localization reduces predictive
errors in the presence of context-dependent model discrepancy.
The results also show why robustness must be assessed relative
to a prescribed context measure. Increasing the localization
rate can improve performance across the full context domain
while degrading performance under a test distribution
concentrated in one regime. Density correction changes this
balance but does not eliminate dependence on calibration context distribution.

In the ocean experiment, amortized localized posterior inference
reduces the aggregate predictive errors and provides
predictive intervals that vary with context. The reference
response is generated by the same numerical model with
context-dependent parameters. Thus, an adequate parameter
exists at each context by construction, while a single global
parameter must represent several forcing regimes. The experiment
therefore tests calibration under the imposed parameter
dependence rather than discrepancy arising from a missing
physical mechanism. The weak-wind heating regime also shows
that local compensability does not ensure parameter
identifiability when the response is insensitive to mixing.

Localization preserves the original numerical model and parameter
interpretation. Its applicability requires local compensability
and sufficient information from nearby contexts. Sparse sampling,
large context uncertainty, or rapid changes in adequate
parameters restrict the useful localization strength.

The conditional flow amortizes inference across contexts,
but flow approximation and optimization remain additional
sources of error.

\section{Conclusion}\label{sec:conclusion}

We constructed localized calibration posteriors using
weighted likelihoods, effective-sample-size scaling, and
a context-noise adjustment. A conditional normalizing flow
approximates the posterior family and allows aggregation
under a prescribed context measure. In the Branin--Hoo twin
experiment, localization reduces aggregate
predictive errors relative to global variational inference.

Localized Bayesian calibration of two $k$--$\varepsilon$
closure coefficients improves temperature
and zonal-velocity predictions in a forcing-dependent
single-column ocean model twin experiment. Both evaluated
localization rates reduce point and ensemble error aggregates
relative to deterministic and global-posterior calibration.
The profiles and marginalized parameter distribution show
adaptation to several imposed forcing regimes, with reduced
parameter identifiability under weak-wind heating conditions.

The method relies on the local structure of the context domain
and the assumption that model discrepancy can be approximately
compensated by calibration parameters.

\FloatBarrier
\clearpage
\appendix
\section{Localized ELBO identity and inference procedures}
\label[appendix]{app:elbo_procedures}

\subsection{Localized ELBO identity}
\label{app:localized_elbo}

For a fixed target context $\CtxStar$, let
\[
\ell_{\CtxStar}(\Param)
:=
\TauLoc{\Bandwidth(\CtxStar)}{\CtxStar}{\Ctxn}
\sum_{i=1}^{n}
\AlgWeighti{i}\log L_i(\Param),
\]
where $L_i$ is the observation likelihood contribution
defined in \cref{sec:localized_posterior}.
The normalized localized posterior is
\begin{equation}\label{eq:localized_posterior_normalized}
\LocalPost(\Param\mid\Dataset_n)
=
\frac{
p_{\rvParam}(\Param)
\exp\{\ell_{\CtxStar}(\Param)\}
}{
\LocalZof{\CtxStar}{\Obsn}{\Ctxn}
},
\end{equation}
with normalization constant
\[
\LocalZof{\CtxStar}{\Obsn}{\Ctxn}
=
\int_{\ParamSpace}
p_{\rvParam}(\Param)
\exp\{\ell_{\CtxStar}(\Param)\}
\,d\Param,
\]
assumed finite and positive.

Substituting this expression into the reverse
Kullback--Leibler divergence gives
\begin{align}\label{eq:elbo_identity_appendix}
\DKL{
\qvarof{\cdot\mid\CtxStar}
}{
\LocalPost(\cdot\mid\Dataset_n)
}
&=
\E_{\qvarof{\cdot\mid\CtxStar}}
\left[
\log\qvarof{\rvParam\mid\CtxStar}
-\log p_{\rvParam}(\rvParam)
-\ell_{\CtxStar}(\rvParam)
\right]
\nonumber\\
&\quad+
\log\LocalZof{\CtxStar}{\Obsn}{\Ctxn}
\nonumber\\
&=
\AlgLocNELBOof{\VarParam}{\CtxStar}
+
\log\LocalZof{\CtxStar}{\Obsn}{\Ctxn},
\end{align}
where $\AlgLocNELBOof{\VarParam}{\CtxStar}$ is the
negative ELBO defined in \cref{eq:localized_nelbo}.
Equivalently,
\[
\LocELBOof{\VarParam}{\CtxStar}{\Obsn}{\Ctxn}
=
-\AlgLocNELBOof{\VarParam}{\CtxStar}
=
\log\LocalZof{\CtxStar}{\Obsn}{\Ctxn}
-
\DKL{
\qvarof{\cdot\mid\CtxStar}
}{
\LocalPost(\cdot\mid\Dataset_n)
}.
\]
Since the Kullback--Leibler divergence is nonnegative,
the localized ELBO is a lower bound on the log
normalization constant. Maximizing it is therefore
equivalent to minimizing the negative ELBO.

\subsection{Localization precomputation}
\label{app:inference_algorithms}

For each target context, the bandwidth is selected to
match the prescribed localization rate $\LocRateTarget$.
An initial upper bracket is obtained from the largest
scaled target--observation distance
\[
d(\CtxStar,\Ctxi{i})
=
\left\|
\CtxScaleMat^{-1}(\CtxStar-\Ctxi{i})
\right\|_2.
\]
The upper bound is expanded until the target rate is
bracketed, after which Brent's method is applied.
The rate is computed using the final normalized weights,
including the selected corrections. If the prescribed
rate cannot be bracketed, a bandwidth cannot be determined
by this procedure.

\begin{algorithm2e}[!htbp]
\caption{Localization precomputation}
\label{alg:localization_precomputation}

\KwIn{Calibration and target contexts;
$\CtxScaleMat$, $\rvCtxNoiseCov$;
target rate $\LocRateTarget$;
lower bandwidth $\hmin$;
bracket factors $\bracketmult>1$, $\bracketexp>1$;
optional $\widehat p_n$ and $\gamma$.}
\KwOut{Bandwidths $\Bandwidth_j$, weights $w_{ji}$,
and scaling factors $\ScalingFactor_j$.}

\For{each target $\Ctx_j^\star$}{
    $\hmax\leftarrow
    \max\{
    \bracketmult\hmin,\,
    \bracketmult\max_i d(\Ctx_j^\star,\Ctxi{i})
    \}$\;

    Expand $\hmax\leftarrow\bracketexp\hmax$
    until the target rate is bracketed, if feasible\;

    Solve
    $\LocRate{\Bandwidth_j}{\Ctx_j^\star}{\Ctxn}
    =\LocRateTarget$
    on $[\hmin,\hmax]$ using Brent's method\;

    Compute the selected kernel weights at
    bandwidth $\Bandwidth_j$, including the
    noise and reflection corrections when applied\;

    Apply the optional density correction and normalize
    to obtain
    $w_{ji}
    =
    \left.\Weighti{i}\right|_{
    \Bandwidth=\Bandwidth_j,\,
    \CtxStar=\Ctx_j^\star
    }$\;

    $\ScalingFactor_j
    \leftarrow
    1/\sum_{i=1}^{n}w_{ji}^2$\;
}
\end{algorithm2e}

\FloatBarrier

\subsection{Conditional-flow training}
\label{app:flow_training}

The conditional flow is trained using the stochastic
negative ELBO in \cref{eq:mcmc_localized_elbo_loss}.
For each target mini-batch, parameter particles are
sampled from the conditional flow, while observation
indices are sampled according to the precomputed
localization weights.

\begin{algorithm2e}[!htbp]
\caption{Stochastic training of the conditional flow}
\label{alg:training}

\KwIn{Calibration data; target contexts;
precomputed $\Bandwidth_j$, $w_{ji}$ and $\ScalingFactor_j$;
number of epochs; target batch size $\btarget$;
observation batch size $\bobs$;
particle count $\nparticle$.}
\KwOut{Trained conditional flow $\qvar$.}

\For{each epoch}{
    Randomly partition target contexts into mini-batches
    of size $\btarget$\;

    \For{each target mini-batch $B$}{
        \For{$j\in B$}{
            Draw $\nparticle$ particles from
            $\qvar(\cdot\mid\Ctx_j^\star)$\;

            Draw $\bobs$ observation indices from
            $\operatorname{Cat}(w_{j1},\ldots,w_{jn})$\;
        }

        Evaluate \cref{eq:mcmc_localized_elbo_loss}
        and differentiate through the reparameterized
        particles\;

        Update $\VarParam$ using the optimizer
        and learning-rate schedule\;
    }
}
\end{algorithm2e}

\FloatBarrier

\clearpage
\section{Localization-kernel illustrations}
\label[appendix]{app:kernel_illustrations}

\subsection{Uniform context sampling}
\label{app:kernel_uniform}

\begin{figure}[!htbp]
  \centering
  \includegraphics[width=370.38235000pt,height=149.56729000pt]{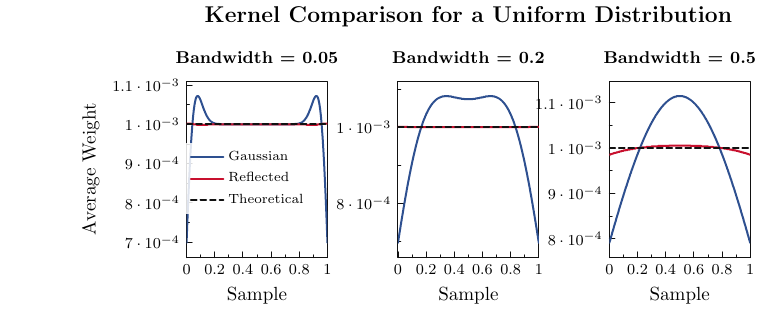}
  \caption{
  Average localization weights across $1000$ observed contexts drawn
  uniformly on $\CtxSpace$. For each observed context value $\Ctxi{i}$
  on the horizontal axis, the plotted quantity is its mean localization
  weight when the target context $\CtxStar$ is also sampled uniformly
  on $\CtxSpace$. The blue curve corresponds to the Gaussian kernel,
  the red curve to the reflected Gaussian kernel, and the dashed
  horizontal line to the nominal uniform weight $1/n$.
  }
  \label{fig:kernel_comparison_uniform}
\end{figure}

\Cref{fig:kernel_comparison_uniform} shows the average
localization weights when the observed contexts and target
contexts are sampled uniformly on $\CtxSpace$.
\FloatBarrier

\subsection{Skewed context sampling}
\label{app:kernel_skewed}

\begin{figure}[!htbp]
  \centering
  \includegraphics[width=370.38235000pt,height=149.57008000pt]{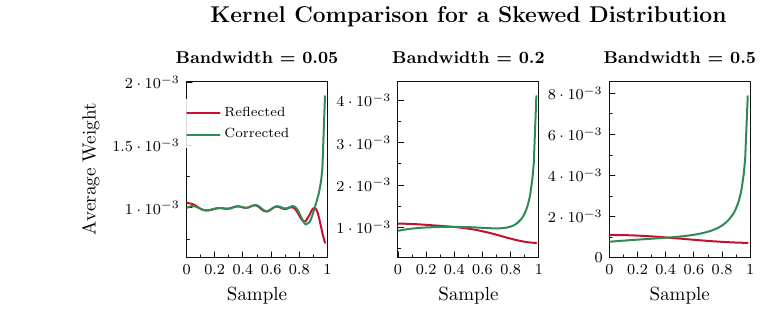}
  \caption{
  Average localization weights across $1000$ observed contexts drawn
  from a skewed distribution on $\CtxSpace$. For each observed context
  value $\Ctxi{i}$, the plotted quantity is its mean localization weight
  when the target context $\CtxStar$ is sampled from the same context
  domain. The red curve corresponds to the reflected Gaussian kernel,
  and the green curve to the reflected kernel with the density correction
  defined in \cref{eq:flattened_localization_weights}.
  }
  \label{fig:kernel_comparison_beta}
\end{figure}

\Cref{fig:kernel_comparison_beta} shows the average
localization weights under skewed context sampling and illustrates
the effect of the optional density correction in
\cref{eq:flattened_localization_weights}.
\FloatBarrier

\clearpage
\section{Branin--Hoo additional results}\label[appendix]{app:branin_results}

\subsection{Point-prediction sensitivity to localization rate}\label{app:effect_of_localization_rate}
\begin{figure}[!htbp]
  \centering
\includegraphics[width=370.39589000pt,height=291.52426000pt]{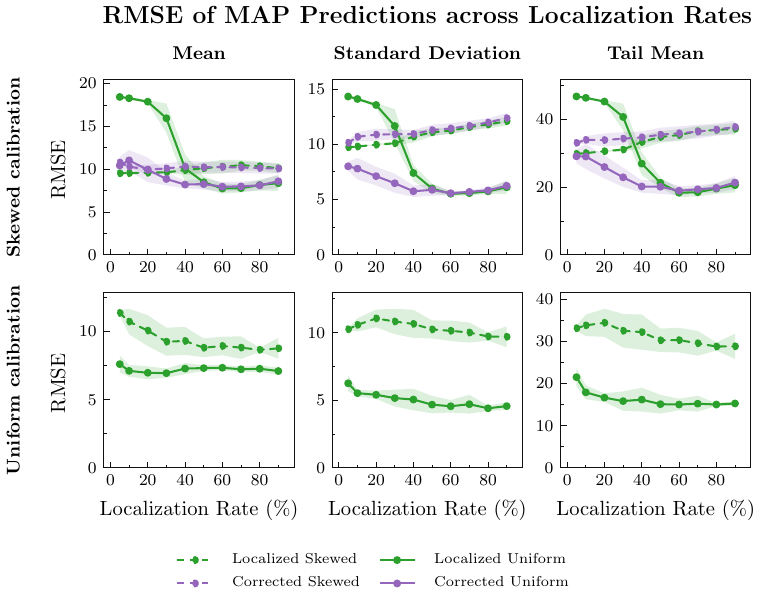}
  \caption{
  Variation of the MAP-based point-prediction error as a function of the localization rate. The first row corresponds to calibration with skewed context data, while the second row corresponds to calibration with uniformly distributed context data.
  }
  \label{fig:map_metrics_per_rate}
\end{figure}
\begin{figure}[!htbp]
  \centering
\includegraphics[width=370.38097000pt,height=267.37698000pt]{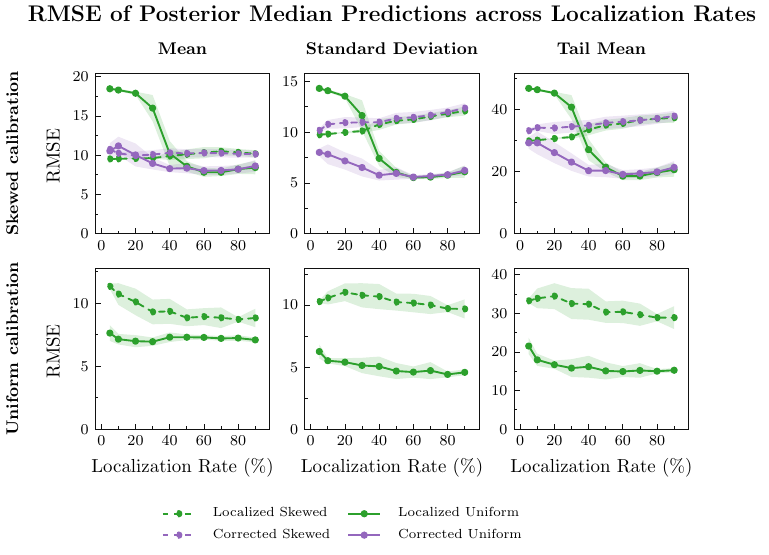}
  \caption{
  Variation of the posterior-median point-prediction error as a function of the localization rate. The first row corresponds to calibration with skewed context data, while the second row corresponds to calibration with uniformly distributed context data.
  }
  \label{fig:median_metrics_per_rate}
\end{figure}
\FloatBarrier

\clearpage
\subsection{Aggregate point-prediction errors}\label{app:branin_point_aggregates}
\begin{table}[!htbp]
\centering
\small
\begin{adjustbox}{max width=\linewidth}
\begin{tabular}{llccc}
\toprule
Method & Summary & $\ErrorMean$ & $\ErrorStd$ & $\ErrorUpperTailMean{0.9}$ \\
\midrule

\multicolumn{5}{l}{\textbf{Uniform calibration; uniform test contexts}} \\
Global
& Median
& \num{44.286} $\pm$ \num{19.106}
& \num{23.318} $\pm$ \num{7.965}
& \num{83.929} $\pm$ \num{29.832} \\
& Mode
& \num{44.288} $\pm$ \num{19.101}
& \num{23.323} $\pm$ \num{7.958}
& \num{83.940} $\pm$ \num{29.813} \\

Localized \num{30}\%
& Median
& \num{6.956} $\pm$ \num{0.332}
& \num{5.136} $\pm$ \num{0.618}
& \num{15.838} $\pm$ \num{2.308} \\
& Mode
& \num{6.952} $\pm$ \num{0.339}
& \num{5.166} $\pm$ \num{0.623}
& \num{15.857} $\pm$ \num{2.298} \\

Localized \num{70}\%
& Median
& \num{7.214} $\pm$ \num{0.188}
& \num{4.723} $\pm$ \num{0.684}
& \num{15.230} $\pm$ \num{1.884} \\
& Mode
& \num{7.234} $\pm$ \num{0.202}
& \num{4.713} $\pm$ \num{0.698}
& \num{15.249} $\pm$ \num{1.879} \\

\midrule
\multicolumn{5}{l}{\textbf{Uniform calibration; skewed test contexts}} \\
Global
& Median
& \num{32.544} $\pm$ \num{13.364}
& \num{22.763} $\pm$ \num{8.544}
& \num{70.723} $\pm$ \num{26.694} \\
& Mode
& \num{32.542} $\pm$ \num{13.366}
& \num{22.758} $\pm$ \num{8.551}
& \num{70.712} $\pm$ \num{26.710} \\

Localized \num{30}\%
& Median
& \num{9.320} $\pm$ \num{0.984}
& \num{10.791} $\pm$ \num{0.976}
& \num{32.570} $\pm$ \num{4.006} \\
& Mode
& \num{9.260} $\pm$ \num{1.027}
& \num{10.829} $\pm$ \num{0.943}
& \num{32.547} $\pm$ \num{3.992} \\

Localized \num{70}\%
& Median
& \num{8.856} $\pm$ \num{0.819}
& \num{10.007} $\pm$ \num{0.728}
& \num{29.679} $\pm$ \num{2.884} \\
& Mode
& \num{8.823} $\pm$ \num{0.830}
& \num{10.014} $\pm$ \num{0.742}
& \num{29.575} $\pm$ \num{2.942} \\

\midrule
\multicolumn{5}{l}{\textbf{Skewed calibration; uniform test contexts}} \\
Global
& Median
& \num{45.673} $\pm$ \num{17.880}
& \num{24.504} $\pm$ \num{6.504}
& \num{87.033} $\pm$ \num{26.146} \\
& Mode
& \num{45.663} $\pm$ \num{17.890}
& \num{24.500} $\pm$ \num{6.507}
& \num{87.018} $\pm$ \num{26.160} \\

Localized \num{30}\%
& Median
& \num{15.997} $\pm$ \num{1.673}
& \num{11.644} $\pm$ \num{1.505}
& \num{40.750} $\pm$ \num{3.869} \\
& Mode
& \num{15.958} $\pm$ \num{1.709}
& \num{11.657} $\pm$ \num{1.529}
& \num{40.750} $\pm$ \num{3.938} \\

Localized corrected \num{30}\%
& Median
& \num{8.950} $\pm$ \num{0.768}
& \num{6.524} $\pm$ \num{0.880}
& \num{23.059} $\pm$ \num{2.828} \\
& Mode
& \num{8.862} $\pm$ \num{0.769}
& \num{6.476} $\pm$ \num{0.861}
& \num{22.898} $\pm$ \num{2.813} \\

Localized \num{70}\%
& Median
& \num{7.820} $\pm$ \num{0.478}
& \num{5.576} $\pm$ \num{0.183}
& \num{18.578} $\pm$ \num{1.322} \\
& Mode
& \num{7.782} $\pm$ \num{0.482}
& \num{5.578} $\pm$ \num{0.196}
& \num{18.497} $\pm$ \num{1.292} \\

Localized corrected \num{70}\%
& Median
& \num{8.058} $\pm$ \num{0.574}
& \num{5.692} $\pm$ \num{0.187}
& \num{19.420} $\pm$ \num{1.346} \\
& Mode
& \num{7.993} $\pm$ \num{0.581}
& \num{5.695} $\pm$ \num{0.164}
& \num{19.359} $\pm$ \num{1.219} \\

\midrule
\multicolumn{5}{l}{\textbf{Skewed calibration; skewed test contexts}} \\
Global
& Median
& \num{31.806} $\pm$ \num{14.588}
& \num{22.689} $\pm$ \num{8.622}
& \num{70.568} $\pm$ \num{27.055} \\
& Mode
& \num{31.807} $\pm$ \num{14.584}
& \num{22.686} $\pm$ \num{8.625}
& \num{70.559} $\pm$ \num{27.067} \\

Localized \num{30}\%
& Median
& \num{9.630} $\pm$ \num{0.082}
& \num{10.123} $\pm$ \num{0.194}
& \num{31.151} $\pm$ \num{0.586} \\
& Mode
& \num{9.639} $\pm$ \num{0.091}
& \num{10.092} $\pm$ \num{0.177}
& \num{31.088} $\pm$ \num{0.559} \\

Localized corrected \num{30}\%
& Median
& \num{10.079} $\pm$ \num{0.560}
& \num{10.977} $\pm$ \num{0.473}
& \num{34.470} $\pm$ \num{1.889} \\
& Mode
& \num{10.091} $\pm$ \num{0.604}
& \num{10.910} $\pm$ \num{0.461}
& \num{34.329} $\pm$ \num{1.899} \\

Localized \num{70}\%
& Median
& \num{10.493} $\pm$ \num{0.517}
& \num{11.555} $\pm$ \num{0.341}
& \num{36.579} $\pm$ \num{1.472} \\
& Mode
& \num{10.466} $\pm$ \num{0.608}
& \num{11.564} $\pm$ \num{0.301}
& \num{36.528} $\pm$ \num{1.528} \\

Localized corrected \num{70}\%
& Median
& \num{10.249} $\pm$ \num{0.520}
& \num{11.676} $\pm$ \num{0.521}
& \num{36.508} $\pm$ \num{1.915} \\
& Mode
& \num{10.195} $\pm$ \num{0.662}
& \num{11.675} $\pm$ \num{0.469}
& \num{36.397} $\pm$ \num{2.061} \\

\bottomrule
\end{tabular}
\end{adjustbox}
\caption{Absolute point-prediction RMSE aggregates for posterior median and mode summaries. Each cell reports the mean $\pm$ one standard deviation over the reported seed runs. The columns report the mean error, standard deviation across test contexts, and upper-tail mean at level $0.9$, respectively.}
\label{tab:point_metrics}
\end{table}
\FloatBarrier

\clearpage
\subsection{Context-wise point errors and relative ensemble diagnostic}\label{app:branin_context_errors}
\begin{figure}[!htbp]
    \centering
\includegraphics[width=368.96407000pt,height=348.86970000pt]{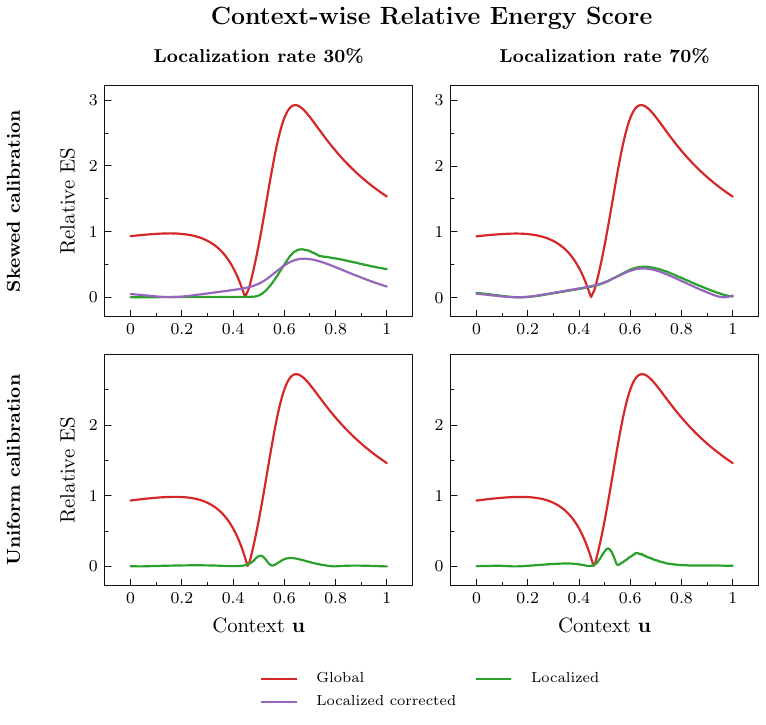}
    \caption{Relative ensemble diagnostic across contexts in the
Branin--Hoo experiment. Rows correspond to skewed and uniform calibration context distributions, while columns compare the selected localization rates.}
    \label{fig:relative_energy_score_per_context}
\end{figure}

\Cref{fig:relative_energy_score_per_context} shows the relative ensemble diagnostic across the context domain. Errors concentrate differently around the designed transition and the less represented regime.
\begin{figure}[!htbp]
    \centering
\includegraphics[width=368.96407000pt,height=347.70096000pt]{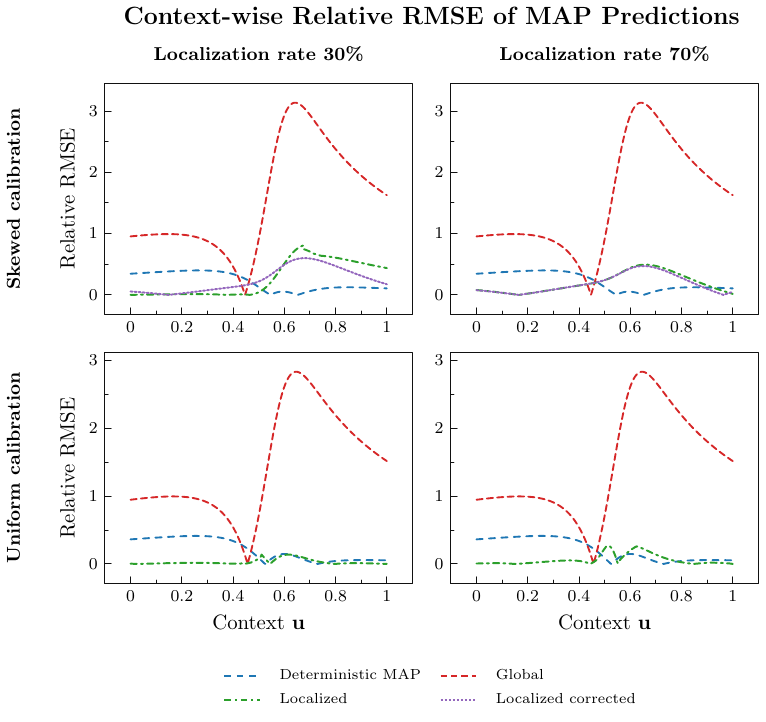}
    \caption{Context-wise MAP point-prediction errors for the Branin--Hoo benchmark. Rows correspond to skewed and uniform calibration context distributions, while columns compare the selected localization rates.}
    \label{fig:errors_per_context_map}
\end{figure}
\begin{figure}[!htbp]
    \centering
\includegraphics[width=370.39252000pt,height=323.35196000pt]{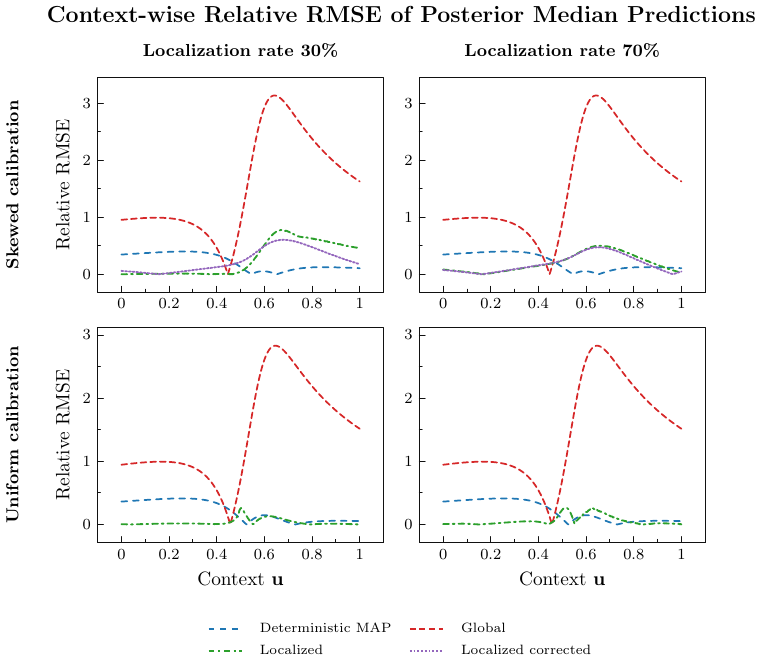}
    \caption{Context-wise posterior-median point-prediction errors for the Branin--Hoo benchmark. Rows correspond to skewed and uniform calibration context distributions, while columns compare the selected localization rates.}
    \label{fig:errors_per_context_median}
\end{figure}
\FloatBarrier

\clearpage
\subsection{Marginalized posteriors at 30\% localization}\label{app:marginalized_posterior}
The marginalized densities at $30\%$ use the same three context measures and calibration distributions as the $70\%$ comparison. The reference lines identify the two regime parameters.
\begin{figure}[!htbp]
    \centering
\includegraphics[width=370.39906000pt,height=301.05850000pt]{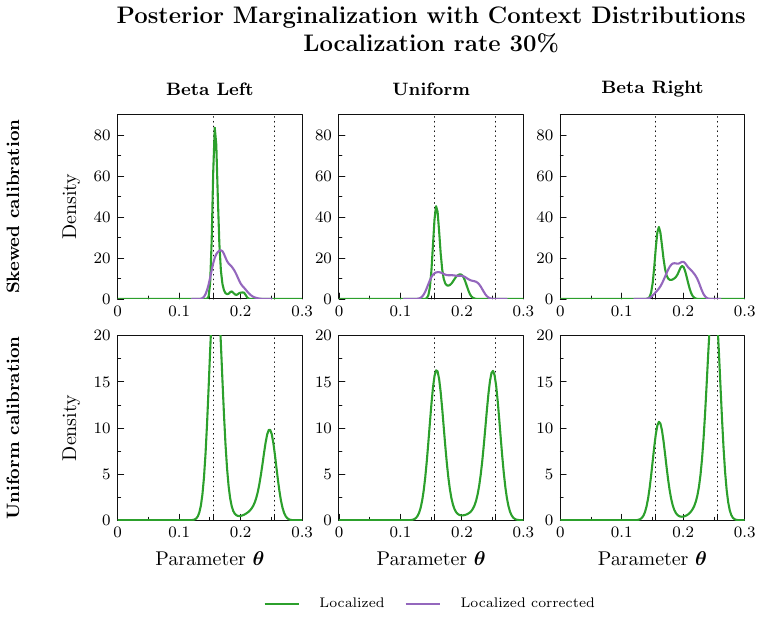}
    \caption{Marginalized posterior distributions for localization rate \num{30}\%, obtained by integrating the localized posterior over different context distributions. The first row corresponds to calibration with skewed context data, while the second row corresponds to calibration with uniform context data. Columns show marginalization with respect to context distributions concentrated toward smaller values, uniform contexts, and distributions concentrated toward larger values. Vertical dotted lines indicate the reference parameter values associated with the two regimes.}
    \label{fig:marginal_posteriors_30}
\end{figure}
\FloatBarrier

\clearpage
\section{Ocean model settings and twin specification}\label[appendix]{app:ocean_model_details}
\subsection{Single-column configuration}\label{app:single_column_configuration}
\begin{table}[!htbp]
\centering
\small
\setlength{\tabcolsep}{5pt}
\renewcommand{\arraystretch}{1.08}
\begin{adjustbox}{max width=\linewidth}
\begin{tabular}{llrl}
\toprule
Parameter & Symbol & Value & Unit \\
\midrule
\multicolumn{4}{l}{\textbf{Grid}} \\
Number of cells       & $\NDepth$       & \num{100}  & -- \\
Total depth           & $\DepthMax$     & \num{2000} & \unit{\metre} \\
Surface control depth & $\DepthControl$ & \num{400}  & \unit{\metre} \\
Stretching parameter  & $\param_{\mathrm{grid}}$        & \num{6.5}  & -- \\

\midrule
\multicolumn{4}{l}{\textbf{Time}} \\
Simulation duration & $\SimDuration$    & \num{72}   & \unit{\hour} \\
Model time step     & $\TimeStep$       & \num{50}   & \unit{\second} \\
Output interval     & $\OutputTimeStep$ & \num{3600} & \unit{\second} \\

\midrule
\multicolumn{4}{l}{\textbf{Physics}} \\
Reference density   & $\RefDensity$        & \num{1027}      & \unit{\kilogram\per\metre\cubed} \\
Gravity             & $\Gravity$           & \num{9.81}      & \unit{\metre\per\second\squared} \\
Heat capacity       & $\HeatCapacity$      & \num{3985}      & \unit{\joule\per\kilogram\per\kelvin} \\
Thermal expansion   & $\ThermalExpansion$  & \num{1.9942e-4} & \unit{\per\kelvin} \\
Haline contraction  & $\HalineContraction$ & \num{8.0e-4}    & \unit{\per\psu} \\
Von Kármán constant & $\VonKarman$         & \num{0.41}      & -- \\
Coriolis parameter  & $\Coriolis$          & \num{0}         & \unit{\per\second} \\

\midrule
\multicolumn{4}{l}{\textbf{Initial conditions}} \\
Surface temperature
& $\Temp_{\mathrm{sfc}}$
& \num{2.0}
& \unit{\degreeCelsius} \\
Temperature gradient
& $\partial \Temp/\partial \Depth$
& \num{1e-3}
& \unit{\kelvin\per\metre} \\
Surface salinity
& $\Sal_{\mathrm{sfc}}$
& \num{32.6}
& \unit{\psu} \\
Salinity gradient
& $\partial \Sal/\partial \Depth$
& \num{0}
& \unit{\psu\per\metre} \\
\bottomrule
\end{tabular}
\end{adjustbox}
\caption{Single-column model configuration parameters.}
\label{tab:scm_config}
\end{table}
\FloatBarrier

\clearpage
\subsection{Turbulence-closure parameter settings}\label{app:turbulence_closure_settings}
\begin{table}[!htbp]
\centering
\small
\begin{adjustbox}{max width=\linewidth}
\begin{tabular}{llr}
\toprule
Group & Symbol & Default value \\
\midrule
\multirow{12}{*}{Stability function}
  & $c_1$    & \num{5.0} \\
  & $c_2$    & \num{0.8} \\
  & $c_3$    & \num{1.968} \\
  & $c_4$    & \num{1.136} \\
  & $c_5$    & \num{0.0} \\
  & $c_6$    & \num{0.4} \\
  & $c_{b1}$ & \num{5.95} \\
  & $c_{b2}$ & \num{0.6} \\
  & $c_{b3}$ & \num{1.0} \\
  & $c_{b4}$ & \num{0.0} \\
  & $c_{b5}$ & \num{0.3333} \\
  & $c_{bb}$ & \num{0.72} \\
\midrule
\multirow{7}{*}{Dissipation and transport}
  & $\ParamCmu$             & \num{0.5477} \\
  & $\sigma_k$              & \num{1.0} \\
  & $\sigma_\varepsilon$    & \num{1.3} \\
  & $\ParamCeps$     & \num{1.44} \\
  & $c_{\varepsilon 2}$     & \num{1.92} \\
  & $c_{\varepsilon 3}^{-}$ & \num{-0.4} \\
  & $c_{\varepsilon 3}^{+}$ & \num{1.0} \\
\midrule
\multirow{3}{*}{GLS exponents}
  & $p$ & \num{3} \\
  & $m$ & \num{1.5} \\
  & $n$ & \num{-1} \\
\bottomrule
\end{tabular}
\end{adjustbox}
\caption{Default parameters of the $k$--$\varepsilon$ turbulence closure.}
\label{tab:kepsilon_default_parameters}
\end{table}
\FloatBarrier

\subsection{Smooth distortion and forcing regimes}\label{app:ocean_distortion}
The smooth transition \cite{mongwane2015toward} is
\begin{equation}\label{eq:twin_smooth_transition}
 \mathcal S_\delta(x)=\begin{cases}
 0,&x\leq\tfrac12-\delta,\\
 6r^5-15r^4+10r^3,&\tfrac12-\delta<x<\tfrac12+\delta,\\
 1,&x\geq\tfrac12+\delta,
 \end{cases}\qquad r=\frac{x-(\tfrac12-\delta)}{2\delta}.
\end{equation}
\begin{figure}[!htbp]
\centering
\includegraphics[width=328.18124000pt,height=157.05759000pt]{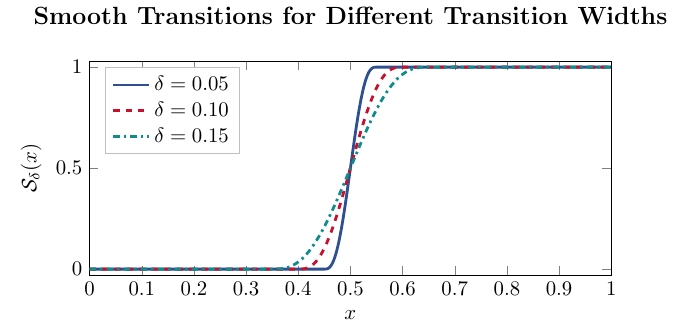}
\caption{Quintic smooth-transition function
        $\mathcal{S}_{\delta}$ for three values of
        $\delta$.}
\label{fig:smoth_quintic}
\end{figure}
\begin{table}[!htbp]
    \centering
    \small
    \setlength{\tabcolsep}{5pt}
    \renewcommand{\arraystretch}{1.25}
    \begin{adjustbox}{max width=\linewidth}
\begin{tabular}{lll}
        \hline
        Regime
        & Plateau conditions
        & $\Distorsion_{\Param}(\Ctx)$
        \\
        \hline

        Low wind and cooling
        &
        $\widebar{\WindForcing}\leq\frac{1}{2}-\delta$,
        $\widebar{\HeatForcing}\leq\frac{1}{2}-\delta$
        &
        $(0,0)$
        \\

        High wind and cooling
        &
        $\widebar{\WindForcing}\geq\frac{1}{2}+\delta$,
        $\widebar{\HeatForcing}\leq\frac{1}{2}-\delta$
        &
        $(0,a_{\varepsilon 1})$
        \\

        Low wind and heating
        &
        $\widebar{\WindForcing}\leq\frac{1}{2}-\delta$,
        $\widebar{\HeatForcing}\geq\frac{1}{2}+\delta$
        &
        $(-a_{\mu 0},0)$
        \\

        High wind and heating
        &
        $\widebar{\WindForcing}\geq\frac{1}{2}+\delta$,
        $\widebar{\HeatForcing}\geq\frac{1}{2}+\delta$
        &
        $(-a_{\mu 0},a_{\varepsilon 1})$
        \\
        \hline
    \end{tabular}
\end{adjustbox}
    \caption{Four main parameter-distortion regimes.}
    \label{tab:twin_distortion_regimes}
\end{table}
\FloatBarrier

\subsection{State--closure coupling and mixed-layer diagnostic}\label{app:scm_diagnostics}
\begin{figure}[!htbp]
\centering
\includegraphics[trim={0.000bp 0.296bp 0.000bp 0.000bp},clip=false,width=370.38814000pt,height=323.41325000pt]{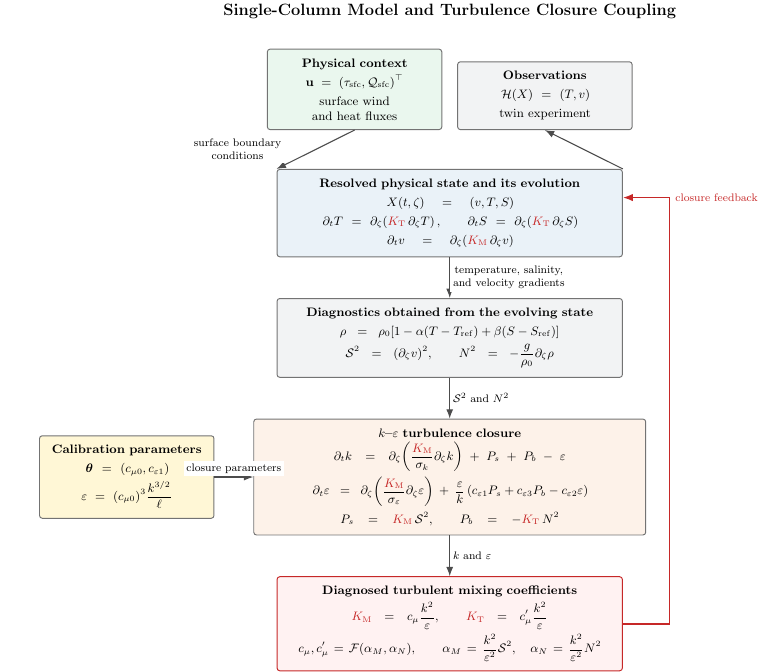}
\caption{Coupling between the physical context, the resolved
single-column state, and the $k$--$\varepsilon$ turbulence closure.
The state determines the local shear and stratification entering the
closure. The closure then returns the turbulent diffusivity
$\DiffVert$ and viscosity $\ViscVert$, shown in red, which close the
state-evolution equations. Temperature and zonal velocity are used as synthetic
observations for calibration.}
\label{fig:scm_closure_coupling}
\end{figure}
The profile figures use the mixed-layer-depth diagnostic of \cite{lemarie2025data}. For the convective configuration, the minimum upward buoyancy flux defines the diagnosed depth. With buoyancy $b=-g(\Density-\RefDensity)/\RefDensity$ and uniform salinity,
\begin{equation}
 \partial_{\Depth}b=g\alpha\partial_{\Depth}\Temp,\qquad F_\Temp^\uparrow=-\DiffVert\partial_{\Depth}\Temp,
 \qquad F_b^\uparrow=g\alpha F_\Temp^\uparrow.
\end{equation}
Since $g\alpha>0$, the two fluxes attain their minimum at the same coordinate:
\begin{equation}\label{eq:scm_mld}
 \MLD(t)=-\operatorname*{arg\,min}_{\Depth\in[-\DepthMax,0]}F_\Temp^\uparrow(t,\Depth).
\end{equation}
The figures use normalized depth $\Depth/\MLD$. The time-averaged profiles are shown on $[-1.2,0]$.
\FloatBarrier

\clearpage
\section{Ocean calibration and training settings}\label[appendix]{app:ocean_calibration_settings}
\subsection{Data and prior}\label{app:ocean_data_settings}
The prior is uniform on $\ParamSpace=[0.3,0.8]\times[0.1,1.9]$ for $(\ParamCmu,\ParamCeps)$.
\begin{table}[!htbp]
    \centering
    \begin{adjustbox}{max width=\linewidth}
\begin{tabularx}{\linewidth}{@{}Xcp{0.38\linewidth}@{}}
        \toprule
        Quantity & Symbol & Value \\
        \midrule
        Calibration data size & $\CalibSize$ & \num{1024} \\
        Test data size & $\TestSize$ & \num{512} \\
        Target contexts & $\TargetSize$ & \num{512} \\
        Temperature noise standard deviation & $\sigma_\Temp$ & $5\%$ of reference-signal std. \\
        Zonal-velocity noise standard deviation & $\sigma_\ZonalVel$ & $5\%$ of reference-signal std. \\
        Wind-forcing noise standard deviation & $\sigma_{\WindForcing}$ &
        \num{4.8e-5} \\
        Heat-forcing noise standard deviation & $\sigma_{\HeatForcing}$ &
        \num{1.5e-5} \\
        Observation noise covariance & $\rvObsNoiseCov$ &
        $\operatorname{diag}\!\left(
            \sigma_\Temp^{2}\mathbf{I}_{m},
            \sigma_\ZonalVel^{2}\mathbf{I}_{m}
        \right)$ \\
        Context noise covariance & $\rvCtxNoiseCov$ &
        $\operatorname{diag}\!\left(
            \sigma_{\WindForcing}^{2},
            \sigma_{\HeatForcing}^{2}
        \right)$ \\
        \bottomrule
    \end{tabularx}
\end{adjustbox}
    \caption{Summary of the calibration configuration.}
    \label{tab:tunax_calibration_config}
\end{table}
\FloatBarrier

\subsection{Optimization, flow architecture, and bandwidth settings}\label{app:ocean_training_settings}
The tables give the data-noise and method-specific
settings. Bandwidth precomputation uses the reflected Gaussian kernel, minimum bandwidth $10^{-6}$, initial bracket distance multiplier 4, expansion factor 2, and at most 100 expansions. A target rate must be attainable and bracketed for root finding.
\begin{table}[!htbp]
\centering
\footnotesize
\setlength{\tabcolsep}{5pt}
\renewcommand{\arraystretch}{0.95}
\begin{adjustbox}{max width=\linewidth}
\begin{tabular}{ll}
\toprule
Setting & Value \\
\midrule
\multicolumn{2}{l}{\textbf{Common training settings}} \\
Number of epochs & \num{45} \\
Initial learning rate & \num{1e-2} \\
Random seed & \num{100} \\

\midrule
\multicolumn{2}{l}{\textbf{Common normalizing-flow settings}} \\
Gradient clipping norm & \num{0.5} \\
Number of flow layers & \num{3} \\
Hidden dimensions & $(\num{64},\num{64},\num{64})$ \\
Number of spline bins & \num{8} \\
Spline bound & \num{5.0} \\

\midrule
\multicolumn{2}{l}{\textbf{MAP baseline}} \\
Batch size & \num{128} \\
Initial value of $\ParamCmu$ & \num{0.65} \\
Initial value of $\ParamCeps$ & \num{1.20} \\

\midrule
\multicolumn{2}{l}{\textbf{Global posterior approximation}} \\
Learning rate decay & \num{0.993443} \\
Batch size & \num{64} \\
Number of particles & \num{16} \\

\midrule
\multicolumn{2}{l}{\textbf{Localized posterior approximation}} \\
Target localization rates & \num{80}\%, \num{98}\% \\
Kernel mode & Reflected Gaussian \\
Context weights & $(\num{1.0},\num{1.0})$ \\
Distance multiplier & \num{4.0} \\
Minimum bandwidth & \num{1e-6} \\
Maximum bracketing expansions & \num{100} \\
Bracketing expansion factor & \num{2.0} \\
Learning rate decay & \num{0.997126} \\
Target-context batch size & \num{16} \\
Number of particles & \num{8} \\
Observation sample size & \num{128} \\
\bottomrule
\end{tabular}
\end{adjustbox}
\caption{Configuration of the MAP baseline, the global normalizing-flow posterior
approximation, and the localized normalizing-flow posterior approximation.}
\label{tab:tunax_baseline_configs}
\end{table}
\FloatBarrier

\clearpage
\section{Final-time ocean profiles}\label[appendix]{app:final_profiles}
The figures show temperature and zonal-velocity
profiles at 72 hours, using the same curves and
posterior bands as the time-averaged comparisons.

\subsection{Temperature}\label{app:final_temperature}
\begin{figure}[!htbp]
    \centering
\includegraphics[width=370.39594000pt,height=384.98552000pt]{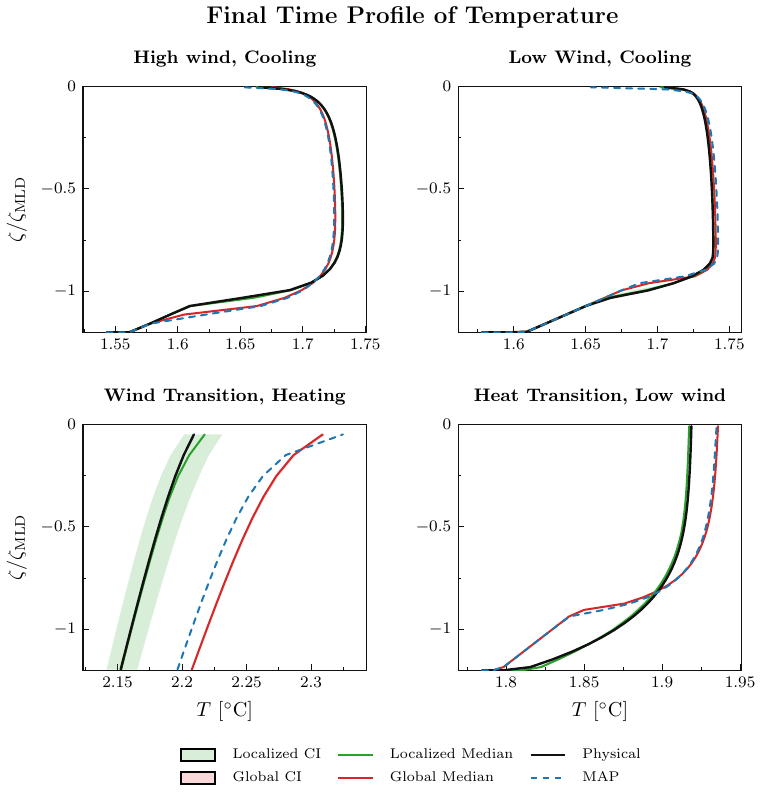}
    \caption{Final temperature profiles after $\qty{72}{\hour}$ as functions of normalized depth $\Depth/\MLD$ for representative forcing regimes. The black curves denote the reference twin-simulation profiles. The green and red curves represent the posterior medians obtained from the localized and global posteriors, respectively, with shaded regions indicating the corresponding $90\%$ credible intervals. The blue dashed curves represent the predictions based on the maximum a posteriori estimates. Localized curves use $98\%$ localization.}
    \label{fig:tunax_last_profile_norm_t}
\end{figure}

\clearpage
\subsection{Zonal velocity}\label{app:final_velocity}
\begin{figure}[!htbp]
    \centering
\includegraphics[width=370.40793000pt,height=395.17836000pt]{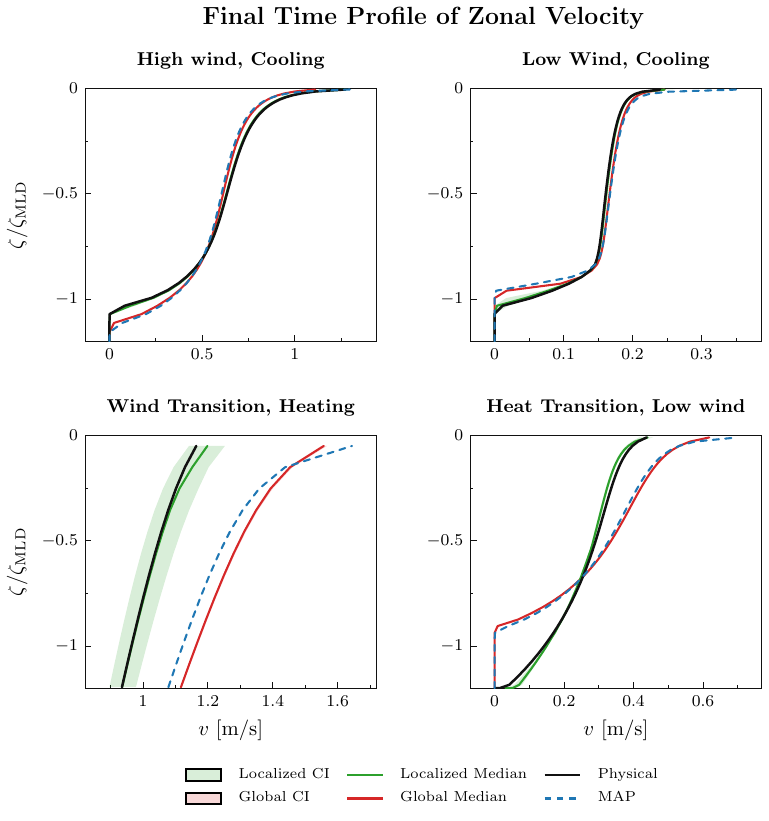}
    \caption{Final zonal-velocity profiles after $\qty{72}{\hour}$ as functions of normalized depth $\Depth/\MLD$ for representative forcing regimes. The black curves denote the reference twin-simulation profiles. The green and red curves represent the posterior medians obtained from the localized and global posteriors, respectively, with shaded regions indicating the corresponding $90\%$ credible intervals. The blue dashed curves represent the predictions based on the maximum a posteriori estimates. Localized curves use $98\%$ localization.}
    \label{fig:tunax_last_profile_norm_u}
\end{figure}
\FloatBarrier

\FloatBarrier
\section*{Acknowledgments}

This work was supported by the French Ministry of
Higher Education and Research through the CIFRE
(Conventions Industrielles de Formation par la Recherche)
program, administered by the Association Nationale de
la Recherche et de la Technologie (ANRT), under
CIFRE agreement No. 2022/1801, in collaboration
with Bull.

This project was provided with HPC/AI computing and
storage resources by GENCI at IDRIS thanks to grant
AD010117241 on the Jean Zay supercomputer's CSL,
A100, and H100 partitions.
\bibliographystyle{siamplain}
\bibliography{references}
\end{document}